\documentclass[10pt,twocolumn,letterpaper]{article}
\pdfoutput=1
\usepackage{iccv}
\usepackage{times}
\usepackage{epsfig}
\usepackage{graphicx}
\usepackage{amsmath}
\usepackage{amssymb}
\usepackage{booktabs} % for better look tables
\usepackage{multirow} % for multirow in tables
\usepackage{siunitx} % for typing \ang{30} as 30 degrees

\usepackage{overpic}
\usepackage{bbm}
\usepackage{dsfont}

\newcommand\boxnet{BoxNet}
\newcommand\votenet{VoteNet}
\newcolumntype{x}[1]{>{\centering\arraybackslash}p{#1pt}}

\usepackage[pagebackref=true,breaklinks=true,letterpaper=true,colorlinks,bookmarks=false]{hyperref}

\iccvfinalcopy % *** Uncomment this line for the final submission

\def\iccvPaperID{1861} % *** Enter the ICCV Paper ID here

\ificcvfinal\fi
\begin{document}

%%%%%%%%% TITLE
\title{Deep Hough Voting for 3D Object Detection in Point Clouds}

\author{%
Charles R. Qi$^{1}$~~~~Or Litany$^{1}$~~~~Kaiming He$^{1}$~~~~Leonidas J. Guibas$^{1,2}$ \vspace{0.2cm}\\
$^{1}$Facebook AI Research~~~~$^{2}$Stanford University
% \vspace{0.1cm}
}

\maketitle
%\thispagestyle{empty}

%%%%%%%%% ABSTRACT
\begin{abstract}
Current 3D object detection methods are heavily influenced by 2D detectors. In order to leverage architectures in 2D detectors, they often convert 3D point clouds to regular grids (i.e., to voxel grids or to bird's eye view images), or rely on detection in 2D images to propose 3D boxes. Few works have attempted to directly detect objects in point clouds.
In this work, we return to first principles to construct a 3D detection pipeline for point cloud data and as generic as possible.
However, due to the sparse nature of the data -- samples from 2D manifolds in 3D space -- we face a major challenge when directly predicting bounding box parameters from scene points: a 3D object centroid can be far from any surface point thus hard to regress accurately in one step.
To address the challenge, we propose \votenet{}, an end-to-end 3D object detection network based on a synergy of deep point set networks and Hough voting.
Our model achieves state-of-the-art 3D detection on two large datasets of real 3D scans, ScanNet and SUN RGB-D with a simple design, compact model size and high efficiency. Remarkably, \votenet{} outperforms previous methods by using purely geometric information without relying on color images.  
\end{abstract}

%%%%%%%%% BODY TEXT
\section{Introduction}
% \begin{itemize}
%     \item 3D object detection is important
%     \item Limitations in current methods (3D CNNs, bird’s eye view detector, frustum-based detector)
%     \item What we want: generic, 3D-based, efficient (leverages sparsity) 3D detector
%     \item Naive way: propose objects on every (subsampled) existing points. Challenges: existing points can be far away from object centers.
%     \item Our approach: deep Hough voting.
%     \item We achieved SOTA results. Key lessons/conclusions.
% \end{itemize}

The goal of 3D object detection is to localize and recognize objects in a 3D scene. More specifically, in this work, we aim to estimate oriented 3D bounding boxes as well as semantic classes of objects from point clouds.
% This is a fundamental task for applications ranging from augmented- and mixed-reality, all the way to home assistant robots and self-driving cars. %In contrast to 2D object detection, the 3D detection problem involves both 3D input and 3D output.
% The input is often a point cloud from either depth cameras or range sensors such as LIDAR.
%

Compared to images, 3D point clouds provide accurate geometry and robustness to illumination changes. On the other hand, point clouds are irregular.
% and inherently sparse, as depth sensors only capture 2D surfaces of objects.
thus typical CNNs are not well suited to process them directly.
% In addition, the output are oriented 3D bounding boxes that have more degrees of freedom compared to the 2D case, which results in a much larger search space. % for 3D object proposals.

\begin{figure}%[h]
    \centering
    \vspace{14pt}
    \begin{overpic}
    [trim=0cm 0cm 0cm 0cm,clip,width=0.95\linewidth]{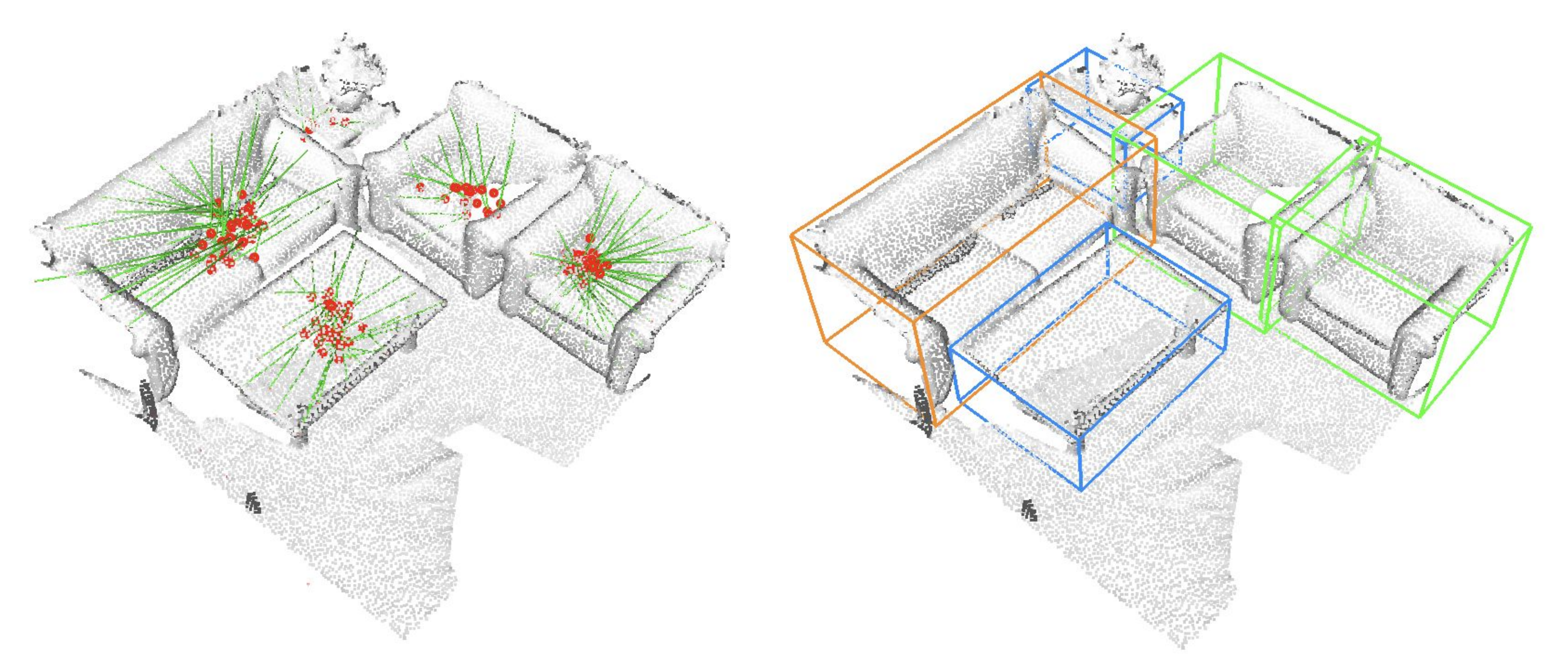}
    \put(2,43){\small Voting from input point cloud}
    \put(60,43){\small 3D detection output}
    \end{overpic}
    \caption{\textbf{3D object detection in point clouds with a deep Hough voting model.} Given a point cloud of a 3D scene, our \votenet{} votes to object centers and then groups and aggregates the votes to predict 3D bounding boxes and semantic classes of objects. Our code is open sourced at \url{https://github.com/facebookresearch/votenet}
    }
    \label{fig:teaser}
\end{figure}

To avoid processing irregular point clouds, current 3D detection methods heavily rely on 2D-based detectors in various aspects. For example, \cite{song2016deep,hou20183d} extend 2D detection frameworks such as the Faster/Mask R-CNN~\cite{ren2015faster,he2017mask} to 3D. They voxelize the irregular point clouds to regular 3D grids and apply 3D CNN detectors, which fails to leverage sparsity in the data and suffer from high computation cost due to expensive 3D convolutions.
% Note that this search is unavoidable in those detectors, because the centroid of 3D objects can be in those empty spaces.
Alternatively,~\cite{cvpr17chen,zhou2018voxelnet} project points to regular 2D bird's eye view images and then apply 2D detectors to localize objects. This, however, sacrifices geometric details which may be critical in cluttered indoor environments. % and robotics use cases.
%which suits well for simple scenes in the driving scenario.
% but loses 3D geometry due to the projection.
% A bird's eye view is suitable in outdoor driving scenes where objects are well-separated and there is rarely an object on top of another--- assumptions that cease to hold in cluttered indoor environments and robotics use cases.
% To avoid a lossy quantization process, a third family of works leverages a 2D view to localize and restrict the 3D search space to a frustum shaped region. These, however, are too dependant on the 2D detector stage and may entirely miss an object if it is not seen by the 2D detector due to occlusion or illumination challenges.
More recently,~\cite{lahoud20172d,qi2018frustum} proposed a cascaded two-step pipeline by firstly detecting objects in front-view images and then localizing objects in frustum point clouds extruded from the 2D boxes, which however is strictly dependent on the 2D detector and will miss an object entirely if it is not detected in 2D.% due to occlusion or hard illumination.
%Both the above two families of work have to convert/rasterize the point clouds before inputting them to CNNs, resulting in information loss from the quantization process. The third family of works is 2D-driven, in the sense that the object is firstly localized in a 2D front-view image and then the 3D detection problem is turned to a 3D localization problem in the frustum point cloud extruded from the 2D object bounding box. This method, however, is too dependent on the 2D detector stage and may entirely miss an object if it is not seen by the 2D detector due to occlusion or illumination challenges.

% So what are the fundamental requirements of a 3D object detector? We desire the method to be (1) \emph{general}, meaning that it is not bound by any assumption on canonical viewpoints (as in the bird's eye view detector); (2) \emph{3D-based}, in the sense that it does not rely on 2D detectors as in 2D-driven methods (while 2D information can still be used by projecting them to 3D); and (3) \emph{efficient}, exploiting sparsity in point clouds and not wasting computation in scanning through large swaths of empty space.

In this work we introduce a \emph{point cloud focused} 3D detection framework that directly processes raw data and does not depend on any 2D detectors neither in architecture nor in object proposal. Our detection network, \emph{\votenet{}}, is based on recent advances in 3D deep learning models for point clouds, and is inspired by the generalized Hough voting process for object detection~\cite{leibe2004combined}.

We leverage PointNet++~\cite{qi2017pointnetplusplus}, a hierarchical deep network for point cloud learning, to mitigates the need to convert point clouds to regular structures. By directly processing point clouds not only do we avoid information loss by a quantization process, but we also take advantage of the sparsity in point clouds by only computing on sensed points.

While PointNet++ has shown success in object classification and semantic segmentation ~\cite{qi2017pointnetplusplus}, few research study how to detect 3D objects in point clouds with such architectures.
%While PointNet++ etc. has proven to perform well for tasks ranging from classification, semantic segmentation etc., little work has investigated how to extend the network for more complicated tasks such as region proposal generation or object detection.
A na\"ive solution would be to follow common practice in 2D detectors and perform dense object proposal~\cite{liu2016ssd,ren2015faster}, i.e. to %directly to the point cloud case, so that we have a backbone point cloud feature learning network and that in turn 
propose 3D bounding boxes directly from the sensed points (with their learned features). %If it is a two-stage scheme, we can also refine the bounding box after RoI (region of interest) pooling of the point features.
However, the inherent sparsity of point clouds makes this approach unfavorable. In images there often exists a pixel near the object center, but it is often not the case in point clouds. As depth sensors only capture surfaces of objects, 3D object centers are likely to be in empty space, far away from any point. As a result, point based networks have difficulty aggregating scene context in the vicinity of object centers. Simply increasing the receptive field does not solve the problem because as the network captures larger context, it also causes more inclusion of nearby objects and clutter.

%much larger offsets need to be regressed. %So is there a better way than directly proposing such object centers?
To this end, we propose to endow point cloud deep networks with a voting mechanism similar to the classical \emph{Hough voting}.  
%which we call \emph{deep Hough voting}.
%
By \emph{voting} we essentially generate new points that lie close to objects centers, which can be \textit{grouped and aggregated} to generate box \textit{proposals}. %See Fig.~\ref{fig:teaser}.

% the proposal problem is decomposed to two stages. This decomposition allows us to focus \emph{attention} on object centers and to more effectively combine information from different parts of the object through a learned \emph{vote pooling}.

In contrast to traditional Hough voting with multiple separate modules that are difficult to optimize jointly, \emph{\votenet{}} is end-to-end optimizable. %All steps, including point cloud feature learning, voting, vote clustering and vote aggregation are implemented through network layers that are fully differentiable. 
Specifically, after passing the input point cloud through a backbone network, we sample a set of seed points and generate votes from their features. Votes are targeted to reach object centers. As a result, vote clusters emerge near object centers and in turn can be aggregated through a learned module to generate box proposals. The result is a powerful 3D object detector that is purely geometric and can be applied directly to point clouds.

%a 3D offset plus an updated feature from the seed. The 3D offset of the vote is directly supervised to point towards the object center, if the seed is on an object. Then the network generates 3D object proposals by finding high density regions of votes and aggregating local votes through a learned vote pooling layer. As local votes may come from different object parts and carry different semantic/geometric information the learned vote pooling plays a key role in combining relevant cues and generating high quality proposals. With the proposals we can either have an extra head to RoI pool the points to refine each proposal and classify the object class, or simply adopt a one-stage detection scheme to directly classify the class along with the proposal step.

We evaluate our approach on two challenging 3D object detection datasets: SUN RGB-D~\cite{song2015sun} and ScanNet~\cite{dai2017scannet}. On both datasets \votenet{}, \emph{using geometry only}, significantly outperforms prior arts that use both RGB and geometry or even multi-view RGB images.
% To deepen the understanding of the importance of voting, we  provide a detailed analysis through comparisons to a network that directly proposes objects from surface points.
% Our study shows that voting effectively broadens the coverage of points from which good proposals are generated, and
Our study shows that the voting scheme supports more effective context aggregation, and verifies that \votenet{}~offers the largest improvements when object centers are far from the object surface (e.g. tables, bathtubs, etc.).
% We also provide analysis experiment to show that voting effectively broadens the coverage of points from which good proposals are generated, and
% confirms that \votenet{}~offers the largest improvement where object centers are far from the object surface (e.g. tables, bathtubs, etc.)

% To get further insight into our proposed model, we have also done sophisticated analysis experiments to understand key modules in the deep Hough voting architecture, including the significance of voting (compared to direct proposals from existing points), the impact of vote clustering, and a validation of the importance of learned vote pooling. We further provide qualitative analysis to understand the strengths and limitations of the method.

In summary, the contributions of our work are:
\begin{itemize}
    \setlength\itemsep{0.1em}
    \item A reformulation of Hough voting in the context of deep learning through an end-to-end differentiable architecture, which we dub \votenet{}.
    \item State-of-the-art 3D object detection performance on SUN RGB-D and ScanNet.
    \item An in-depth analysis of the importance of voting for 3D object detection in point clouds.
\end{itemize}

\begin{figure*}[t!]
    \centering
    \includegraphics[width=0.95\linewidth]{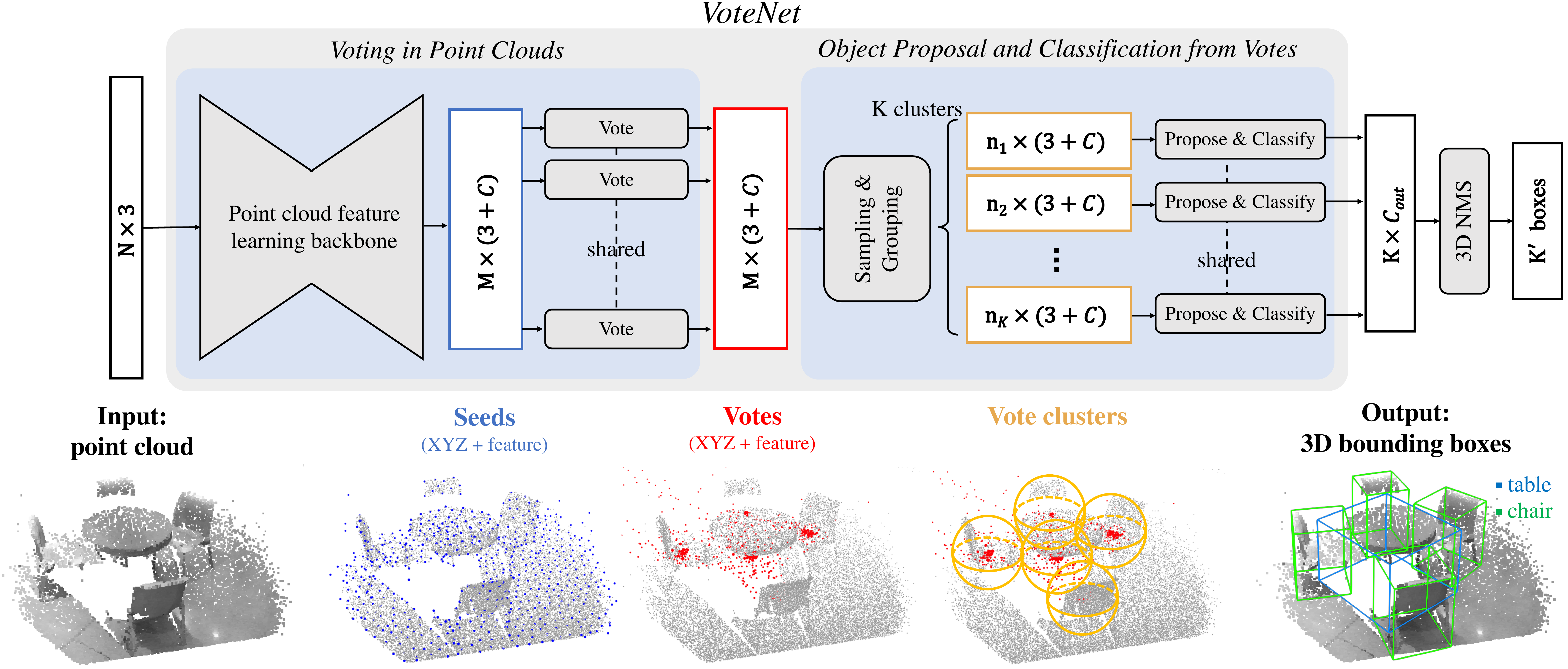}
    \caption{\textbf{Illustration of the \votenet~architecture} for 3D object detection in point clouds. Given an input point cloud of $N$ points with XYZ coordinates, a backbone network (implemented with PointNet++~\cite{qi2017pointnetplusplus} layers) subsamples and learns deep features on the points and outputs a subset of $M$ points but extended by $C$-dim features. This subset of points are considered as seed points. Each seed independently generates a vote through a voting module. Then the votes are grouped into clusters and processed by the proposal module to generate the final proposals. The classified and NMSed proposals become the final 3D bounding boxes output. Image best viewed in color.}
    \label{fig:votingnet}
\end{figure*}

\section{Related Work}
\paragraph{3D object detection.}
% is a long-standing problem in computer vision.
% Traditionally, the term detection has been overloaded and used to describe many related tasks such as registration of an existing set of instances, and object recognition to name a few.
% In this work we refer to 3D object detection as the task of locating and labeling objects belonging to a pre-defined set of semantic classes in an input 3D scene.
Many previous methods were proposed to detect 3D bounding boxes of objects.
Examples include: \cite{lin2013holistic} where a pair-wise semantic context potential helps guide the proposals objectness score;  template-based methods~\cite{li2015database, Indoor2012, litany2015asist}; Sliding-shapes \cite{song2014sliding} and its deep learning-based successor \cite{song2016deep}; Clouds of Oriented Gradients (COG)~\cite{ren2016three}; and the recent 3D-SIS \cite{hou20183d}.
% As some successful methods used 3D CNNs they are computationally heavy. One way to circumvent the complexity is to exploit sparsity of point cloud data, as was recently proposed by GSPN~\cite{yi2018gspn} and PointRCNN~\cite{shi2018pointrcnn} in concurrent to our work.

Due to the complexity of directly working in 3D, especially in large scenes, many methods resort to some type of projection.   %(especially for scenes in driving scenarios). 
For example in MV3D~\cite{cvpr17chen} and VoxelNet \cite{zhou2018voxelnet}, the 3D data is first reduced to a bird's-eye view before proceeding to the rest of the pipeline. A reduction in search space by first processing a 2D input was demonstrated in both Frustum PointNets \cite{qi2018frustum} and \cite{lahoud20172d}. Similarly, in \cite{kim2013accurate} a segmentation hypothesis is verified using the 3D map. 
More recently, deep networks on point clouds are used to exploit sparsity of the data by GSPN~\cite{yi2018gspn} and PointRCNN~\cite{shi2018pointrcnn}.% which are in concurrent to our work.

% \orl{Discuss complex YOLO: \cite{simon2018complex} ?}

\paragraph{Hough voting for object detection.}
Originally introduced in the late 1950s, the Hough transform \cite{hough1959machine} translates the problem of detecting simple patterns in point samples to detecting peaks in a parametric space. The Generalized Hough Transform~\cite{ballard1981generalizing} further extends this technique to image patches as indicators for the existence of a complex object. Examples of using Hough voting include the seminal work of \cite{leibe2008robust} which introduced the implicit shape model, planes extraction from 3D point clouds~\cite{borrmann20113d}, and 6D pose estimation~\cite{sun2010depth} to name a few. 
% This technique was extremely popular in object detection literature until a decade ago, including the
% \cite{leibe2008robust} is a seminal work along this line that introduced the implicit shape model (ISM). detection of planes in 3D point clouds~\cite{borrmann20113d}, Multiple object detection~\cite{barinova2012detection}, and 6D pose estimation~\cite{sun2010depth} to name a few.  

Hough voting has also been previously combined with advanced learning techniques. In ~\cite{maji2009object} the votes were assigned with weights indicating their importance, which were learned using a max-margin framework.  
% As we will describe in Section, in our method we also use supervision to better separate foreground and background votes.
Hough forests for object detection were introduced in~\cite{gall2011hough,gall2013class}. More recently, \cite{kehl2016deep} demonstrated improved voting-based 6D pose estimation by using deep features extracted to build a codebook. Similarly \cite{milletari2017hough} learned deep features to build codebooks for segmentation in MRI and ultrasiounds images. In~\cite{huan2017vehicle} the classical Hough algorithm was used to extract circular patterns in car logos, which were then input to a deep classification network.
% A 2D voting-based part-detection network was recently introduced, utilizing votes from observable parts to predict occluded ones. This relates to our method in that we predict object centers which are often not part of the object itself.
\cite{novotny2018semi} proposed the semi-convolutional operator for 2D instance segmentation in images, which is also related to Hough voting.

There have also been works using Hough voting for 3D object detection~\cite{woodford2014demisting,knopp2010orientation,velizhev2012implicit,knopp2011scene}, which adopted a similar pipeline as in 2D detectors.
%So they retain the same limitations as those for the 2D versions.

% \todo{Start another paragraph to specifically talk about methods that try to use deep networks for hough voting and discuss the differences between our method and theirs.}

% DeepVoting: A Robust and Explainable Deep Network for Semantic Part Detection under Partial Occlusion. CVPR 2018 from Alan Y.~\cite{zhang2018deepvoting}

% Deep Learning of Local RGB-D Patches for 3D Object Detection and 6D Pose Estimation ECCV 2016~\cite{kehl2016deep}

% Depth-Encoded Hough Voting for Joint Object
% Detection and Shape Recovery ECCV 2010~\cite{sun2010depth}

% Deep residual Hough voting for mitotic cell detection in histopathology images~\cite{wollmann2017deep}

% Hough-CNN~\cite{milletari2017hough}

% ICCV 2017 Hough voting~\cite{huan2017vehicle} for Vehicle Logo Retrieval

% Semi-convolutional operator ECCV 2018~\cite{novotny2018semi}

% Hough forest (trained random forest to map from image patch to votes) ~\cite{gall2011hough,gall2013class}

% Explain that vote in the work of \cite{engelcke2017vote3deep} means something entirely different. 

\paragraph{Deep learning on point clouds.}
% Out of the many 3D representations available, point clouds seem to offer a powerful functionality while being memory efficient and simple to store. Moreover, they are the raw output of many depth sensors, and thus can be thought of the 3D equivalent of image pixels as they represent \textit{what is}, as opposed to voxel which also represent \textit{what is not}.
Recently we see a surge of interest in designing deep network architectures suited for point clouds~\cite{qi2017pointnet, qi2017pointnetplusplus, su2018splatnet, atzmon2018point, li2018pointcnn, graham20183d, wang2017cnn, tatarchenko2017octree,tatarchenko2018tangent,le2018pointgrid,klokov2017escape,yang2018foldingnet,xu2018spidercnn,wang2018dynamic,xie2018attentional}, which showed remarkable performance in 3D object classification, object part segmentation, as well as scene segmentation. In the context of 3D object detection, VoxelNet~\cite{zhou2018voxelnet} learn voxel feature embeddings from points in voxels, while in~\cite{qi2018frustum} PointNets are used to localize objects in a frustum point cloud extruded from a 2D bounding box. However, few methods studied how to directly propose and detect 3D objects in raw point cloud representation.
% Recently an efficient sparse convolution network~\cite{graham20183d} was shown to achieve impressive performance on 3D semantic segmentation. In this work we use PointNet++ as our backbone architecture. 

% Starts from 3d shape classification, goes to semantic segmentation, recently used as a module in 3D object detection (voxelnet, frustum pointnets) but not yet fully utilized.

%pointnet~\cite{qi2017pointnet}, 

% pointnet++~\cite{qi2017pointnetplusplus}, SPLATNet~\cite{su2018splatnet}
% SubmanifoldSparseConv~\cite{graham20183d}
% PointCNN~\cite{li2018pointcnn}

\section{Deep Hough Voting}
\label{sec:deep_hough_voting}
% As our pipeline is based on the traditional Hough voting procedure, for completeness we provide a brief description of this well known technique.
%For more detailed explanation, we refer the reader to \orl{TBD}.

A traditional Hough voting 2D detector~\cite{leibe2008robust} comprises an offline and an online step. First, given a collection of images with annotated object bounding boxes, a codebook is constructed with stored mappings between image patches (or their features) and their offsets to the corresponding object centers.  
%
%involves multiple separate modules and steps in both offline processing and online inference. At offline time, one needs to first construct a codebook that stores mappings from image patches (or their features) to votes (as 3D offset vectors from the pixels to object centers). 
%
At inference time, interest points are selected from the image to extract patches around them. These patches are then compared against patches in the codebook to retrieve offsets and compute votes. As \textit{object} patches will tend to vote in agreement, clusters will form near object centers. Finally, the object boundaries are retrieved by tracing cluster votes back to their generating patches. 

%and then match patches around interest points to the patches in the codebook. The votes in the codebook are then copied to the query image to create votes. It happens that a clutter image patch can also create votes, but it is fine as long as they do not agree with each other on where to vote (i.e. they do not create vote clusters). Next, we need to cluster the generated votes to find peaks, which are potential locations of object centers. To recover the object shape, the votes are back-projected to their original image patches, where a bounding box can be computed to cover those patches.

We identify two ways in which this technique is well suited to our problem of interest.
% First, voting is designed for sparse sets, and hence is a natural fit to point clouds.
First, voting-based detection is more compatible with sparse sets than region-proposal networks (RPN)~\cite{ren2015faster} is. For the latter, the RPN has to generate a proposal near an object center which is likely to be in an empty space, causing extra computation.
Second, it is based on a bottom-up principle where small bits of partial information are accumulated to form a confident detection. Even though neural networks can potentially aggregate context from a large receptive field, it may be still beneficial to aggregate in the vote space. %object scans are often very partial, which may lead to ambiguous decisions if not aggregated properly.
% Even more crucial is the lack of object points near the centroid which prohibits direct context aggregation. 

%One strength of the Hough voting detection pipeline is it only computes votes on interest points without dense computation on all pixels. So the method is well suited to sparse data such as a point cloud, where we hope not to compute at every spatial location (including the empty ones) but rather just want to compute on existing points on the object surfaces. Other strengths of hough voting including its flexibility to make use of object templates and robustness to occlusions [add refs].

However, as traditional Hough voting comprises multiple separated modules, integrating it into state-of-the-art point cloud networks is an open research topic. To this end, we propose the following adaptations to the different pipeline ingredients.
\smallskip

%
% for detection is that it contains multiple separated modules each with a handful of parameters thus the whole system is not end-to-end optimizable.
%

% a new scheme of Hough voting in the era of deep learning. Based on a point cloud deep network backbone, our proposed \votenet{} end-to-end optimizable. 

% Below we show how a deep hough voting system aligns/compares with a traditional hough voting pileine.

% \begin{itemize}

\noindent \textbf{Interest points}
% can be learned to be identified using object point supervision instead of hand-crafted features.
are described and selected by deep neural networks instead of depending on hand-crafted features.

\smallskip
    %
    % Instead of using heuristics or hand-crafted features to select interest points, a deep network can learn to select and sample interest points.
    %
    
\noindent\textbf{Vote} generation is learned by a network instead of using a codebook. Levaraging larger receptive fields, voting can be made less ambiguous and thus more effective. In addition, a vote \emph{location} can be augmented with a \emph{feature vector} allowing for better aggregation. 

%    Instead of relying on a codebook, we can learn to map from point features to votes with neural networks. In addition, rather than relying on local patches, we can learn deep point features with large receptive field such that lots of ambiguities can be resolved. Compared with traditional voting, now we can vote both much more efficiently (no need for k-NN search in a large codebook) but also more effectively (much higher vote quality due to increased context through learning).
%

\smallskip
\noindent\textbf{Vote aggregation} is realized through point cloud processing layers with trainable parameters. Utilizing the vote features, the network can potentially filter out low quality votes and generate improved proposals.

% A network can evaluate and generate cluster scores for all local neighborhoods of votes in parallel, and even learn how to aggregate votes to generate better proposals. A clustering step (which often does not gradient descent) can be totally eliminated in the pipeline. The aggregation step can further filter out low quality votes from clutters and combine complementary cues of votes from different parts of a object.

\smallskip
\noindent\textbf{Object proposals} in the form of: location, dimensions, orientation and even semantic classes can be directly generated from the aggregated features, mitigating the need to trace back votes' origins.

% From the aggregated votes, we can estimate the object scale, pose, class and refine the object location from the vote features. There is no need to back-project from votes to original patches.

% \end{itemize}
In what follows, we describe how to combine all the aforementioned ingredients into a single end-to-end trainable network named as \votenet{}.

% The described end-to-end trainable pipeline 

% With an end-to-end trained point cloud network, every step in the pipeline can be optimized through gradient descent for the final detection goal. In the next section, we will describe an instantiation of the deep hough voting scheme, a 3D object detection network called the VotingNet.

\section{VoteNet Architecture}
% In this section, we describe our proposed deep hough voting network (named \emph{VotingNet}) for 3D object detection in point clouds.

Fig.~\ref{fig:votingnet} illustrates our end-to-end detection network (\emph{\votenet{}}).
% learns to vote to object centers from the input point cloud and to propose and classify 3D objects from the votes.
The entire network can be split into two parts: one that processes \emph{existing} points to generate votes; and the other part that operates on \emph{virtual} points -- the votes -- to propose and classify objects.

\subsection{Learning to Vote in Point Clouds}
\label{sec:votingnet:vote}
% \rqi{Should we mention somewhere the entire network is a large PointNet++ network except that we have an intermediate voting supervision and switched from learning on existing points to votes after the voting layer.}

% Input to our network is a point cloud of a scene in size $N \times 3$ with $N$ points and each point has its 3D coordinates. In the voting step we aim to generate a set of \emph{votes}  in size $M \times F \times (3+C_2)$ from a subsampled set of points (seeds) in size $M \times (3+C_1)$, where $F$ is the voting factor (how many votes we generate from each seed point).
From an input point cloud of size $N \times 3$, with a 3D coordinate for each of the $N$ points, we aim to generate $M$ votes, where each vote has both a 3D coordinate and a high dimensional feature vector. There are two major steps: point cloud feature learning through a backbone network and learned Hough voting from seed points.

% A voting process involves three key definitions:
% \begin{itemize}
%     \item Seeds: These are the ``voters'' and are where the votes come from. They are also referred to as interest points.
%     \item Votes: a vote is a signal generated from a seed that indicates the geometry and semantics of the object of interest.
% \end{itemize}

% Compared to traditional generalized hough transform, voting in our method is different in two aspects. First, the VotingNet does not rely on a codebook that records the mappings from local patch features to votes. Instead, the network learns to generate votes from point cloud features with a trained neural network. Second, the representation of votes in our network is no longer just a 3D offset but rather a 3D offset plus a high-dimensional vote feature vector.

\smallskip\noindent\textbf{Point cloud feature learning.}
% \paragraph{Point cloud feature learning.}
Generating an accurate vote requires geometric reasoning and contexts. Instead of relying on hand-crafted features, we leverage recently proposed deep networks~\cite{qi2017pointnetplusplus,graham20183d,su2018splatnet,li2018pointcnn} on point clouds for point feature learning. While our method is not restricted to any point cloud network, we adopt PointNet++~\cite{qi2017pointnetplusplus} as our backbone due to its simplicity and demonstrated success on tasks ranging from normal estimation~\cite{guerrero2018pcpnet}, semantic segmentation~\cite{landrieu2018large} to 3D object localization~\cite{qi2018frustum}.

The backbone network has several set-abstraction layers and feature propagation (upsampling) layers with skip connections, which outputs a subset of the input points with XYZ and an enriched $C$-dimensional feature vector. The results are $M$ \emph{seed points} of dimension $(3+C)$. Each seed point generates one vote\footnote{The case of more than one vote is discussed in the appendix.}.

\smallskip\noindent\textbf{Hough voting with deep networks.} Compared to traditional Hough voting where the votes (offsets from local keypoints) are determined by look ups in a pre-computed codebook, we generate votes with a deep network based voting module, which is both more efficient (without kNN look ups) and more accurate as it is trained jointly with the rest of the pipeline.

Given a set of seed points $\{s_i\}_{i=1}^{M}$ where $s_i = [x_i; f_i]$ with $x_i \in \mathbb{R}^3$ and $f_i \in \mathbb{R}^{C}$, a shared \emph{voting module} generates votes from each seed independently. Specifically, the voting module is realized with a multi-layer perceptron (MLP) network with fully connected layers, ReLU and batch normalization. The MLP takes seed feature $f_i$ and outputs the Euclidean space offset $\Delta x_i \in \mathbb{R}^3$ and a feature offset $\Delta f_i \in \mathbb{R}^{C}$ such that the vote $v_i = [y_i; g_i]$ generated from the seed $s_i$ has $y_i = x_i + \Delta x_i$ and $g_i = f_i + \Delta f_i$.

The predicted 3D offset $\Delta x_i$ is explicitly supervised by a regression loss
\begin{equation}
    L_{\text{vote-reg}} = \frac{1}{M_{\text{pos}}} \sum_i  \|\Delta x_i - \Delta x_i^*\| \mathds{1}[s_i \text{ on object}], 
\end{equation}

\noindent
where $\mathds{1}[s_i \text{ on object}]$ indicates whether a seed point $s_i$ is on an object surface and $M_\text{pos}$ is the count of total number of seeds on object surface. $\Delta x_i^*$ is the ground truth displacement from the seed position $x_i$ to the bounding box center of the object it belongs to. 

% From each seed point, the voting layer generates $F$ votes from the seed point's feature, where $F$ is the voting factor (in default $F=1$). Each vote consists of a 3D offset relative to the seed point's 3D location and a residual feature prediction relative to the seed feature. The vote features can carry information of where the vote is from and also carry any useful semantic or geometric information useful for the later object proposal step.

Votes are the same as seeds in tensor representation but are no longer grounded on object surfaces. A more essential difference though is their position -- votes generated from seeds on the same object are now closer to each other than the seeds are, which makes it easier to combine cues from different parts of the object. Next we will take advantage of this semantic-aware locality to aggregate vote features for object proposal.

\subsection{Object Proposal and Classification from Votes}
\label{sec:votingnet:pooling}

% Traditionally in generalized hough voting, after the voting step, peaks of votes are found through some clustering algorithm and then the votes in high density clusters are back-projected to their original interest points/patches to recover objects. The clustering and peak search process is necessary because they can reduce the number of proposals and only keep those with high confidence i.e. large density. However this process involves many parameters such as choosing grid size, density thresholds and criterion for convergence on clustering, and most importantly not friendly to gradient descent.

The votes create canonical ``meeting points'' for context aggregation from different parts of the objects. After clustering these votes we aggregate their features to generate object proposals and classify them. 

\smallskip\noindent\textbf{Vote clustering through sampling and grouping.}
While there can be many ways to cluster the votes, we opt for a simple strategy of uniform sampling and grouping according to spatial proximity. Specifically, from a set of votes $\{v_i\ = [y_i; g_i] \in \mathbb{R}^{3+C}\}_{i=1}^{M}$, %where $v_i = [y_i; g_i]$ with $y_i \in \mathbb{R}^3$ and $g_i \in \mathbb{R}^{C}$
we sample a subset of $K$ votes using farthest point sampling based on $\{y_i\}$ in 3D Euclidean space, to get $\{v_{i_k}\}$ with $k=1,...,K$. Then we form $K$ clusters by finding neighboring votes to each of the $v_{i_k}$'s 3D location: $\mathcal{C}_k = \{v_i^{(k)} | \| v_i - v_{i_k}\| \leq r\}$ for $k=1,...,K$. Though simple, this clustering technique is easy to integrate into an end-to-end pipeline and works well in practice.

\smallskip\noindent\textbf{Proposal and classification from vote clusters.}
As a vote cluster is in essence a set of high-dim points, we can leverage a generic point set learning network to aggregate the votes in order to generate object proposals. Compared to the back-tracing step of traditional Hough voting for identifying the object boundary, this procedure allows to propose \textit{amodal} boundaries even from partial observations, as well as predicting other parameters like orientation, class, etc.  

In our implementation, we use a \emph{shared} PointNet~\cite{qi2017pointnet} for vote aggregation and proposal in clusters. Given a vote cluster $\mathcal{C} = \{w_i\}$ with $i=1,...,n$ and its cluster center $w_j$, where $w_i = [z_i; h_i]$ with $z_i \in \mathbb{R}^3$ as the vote location and $h_i \in \mathbb{R}^{C}$ as the vote feature. To enable usage of local vote geometry, we transform vote locations to a local normalized coordinate system by $z'_i = (z_i - z_j)/r$. Then an object proposal for this cluster $p(\mathcal{C})$ is generated by passing the set input through a PointNet-like module:

\begin{equation}
    p(\mathcal{C}) = \text{MLP}_2 \left\{ \underset{i=1,...,n}{\mbox{max}}\left\{\text{MLP}_1 ([z'_i; h_i])\right\} \right\}
    \label{eq:vote_aggregation}
\end{equation}

\noindent
where votes from each cluster are independently processed by a $\text{MLP}_1$ before being max-pooled (channel-wise) to a single feature vector and passed to $\text{MLP}_2$ where information from different votes are further combined. We represent the proposal $p$ as a multidimensional vector with an objectness score, bounding box parameters (center, heading and scale parameterized as in~\cite{qi2018frustum}) and semantic classification scores.

\smallskip\noindent\textbf{Loss function.}
% \todo{Make everything consistent with the notations.}
% The network is supervised by a multi-task loss (Eq.~\ref{eq:loss}) including a voting regression loss ($L1$), an object proposal loss (cross entropy) and a object detection loss for bounding box parameter estimation and semantic classification.
The loss functions in the proposal and classification stage consist of objectness, bounding box estimation, and semantic classification losses.

% The voting is supervised for its $XYZ$ offset. In SUN RGB-D, since there is no annotations for 3D instance segmentation, we consider all points in an object's bounding box should vote for the object center. To deal with cases a point lives in multiple object's bounding boxes (e.g. a chair point under the table), we assign multiple (three in our implementation) ground truth vote labels to a point and evaluate a minimum of $K$ loss when comparing a predicted vote offset with the set of ground truth votes. In ScanNetV2, points have instance labels so we can acquire a one-to-one mapping of object point to vote label. However, bounding boxes in ScanNetV2 is not amodal, so we just calculate the bounding box of the visible points and compute the ground truth vote as the one points to the visible box center.

We supervise the objectness scores for votes that are located either close to a ground truth object center (within $0.3$ meters) or far from any center (by more than $0.6$ meters). We consider proposals generated from those votes as \textit{positive} and \textit{negative} proposals, respectively. Objectness predictions for other proposals are not penalized. Objectness is supervised via a cross entropy loss normalized by the number of non-ignored proposals in the batch. For positive proposals we further supervise the bounding box estimation and class prediction according to the closest ground truth bounding box. Specifically, we follow ~\cite{qi2018frustum} which decouples the box loss to center regression, heading angle estimation and box size estimation. For semantic classification we use the standard cross entropy loss. In all regression in the detection loss we use the Huber (smooth-$L_1$~\cite{ren2015faster}) loss. Further details are provided in the appendix.

\subsection{Implementation Details}
\label{sec:votingnet:implmentation}

\smallskip\noindent\textbf{Input and data augmentation.} Input to our detection network is a point cloud of $N$ points randomly sub-sampled from either a popped-up depth image ($N=20k$) or a 3D scan (mesh vertices, $N=40k$).
In addition to $XYZ$ coordinates, we also include a height feature for each point indicating its distance to the floor. The floor height is estimated as the $1\%$ percentile of all points' heights.
%Note that since our network is translation invariant, it does not use any absolute coordinate of the points so the height channel is necessary. %So the final input is a point cloud of size $20,000 \times 4$.
To augment the training data, we randomly sub-sample the points from the scene points on-the-fly. We also randomly flip the point cloud in both horizontal direction, randomly rotate the scene points by $\text{Uniform}[-5^{\circ},5^{\circ}]$ around the upright-axis, and randomly scale the points by $\text{Uniform}[0.9,1.1]$.

\smallskip\noindent\textbf{Network architecture details.} The backbone feature learning network is based on PointNet++~\cite{qi2017pointnetplusplus}, which has four set abstraction (SA) layers and two feature propagation/upsamplng (FP) layers, where the SA layers have increasing receptive radius of $0.2$, $0.4$, $0.8$ and $1.2$ in meters while they sub-sample the input to $2048$, $1024$, $512$ and $256$ points respectively. The two FP layers up-sample the 4th SA layer's output back to $1024$ points with $256$-dim features and 3D coordinates (more details in the appendix).

The voting layer is realized through a multi-layer perceptron with FC output sizes of $256,256,259$, where the last FC layer outputs XYZ offset and feature residuals.
% The voting layer involves a vote regression FC layer and a feature update FC layer shared across all seed points. For a seed point $s_i$ with 3D coordinate $X(s_i)$ and feature $F(s_i)$. The output vote $v_i$'s coordinate is $X(v_i) = FC_1(F(s_i)) + X(s_i)$ and its feature is $F(v_i) = FC_2(F(s_i) + F(s_i)$, where $FC_1$ maps from 256-dim feature to 3-dim offset and $FC_2$ maps 256-dim feature to 256-dim feature residual. Both layers do not have non-linear or BatchNorm.

% The vote cluster is sampled with farthest point sampling on the votes during training. At inference time, different sampling strategies can be taken as we will discuss more in the experiment section. $256$ proposals are generated for each scene.
The proposal module is implemented as a set abstraction layer with a post processing $\text{MLP}_2$ to generate proposals after the max-pooling. The SA uses radius $0.3$ and $\text{MLP}_1$ with output sizes of $128,128,128$. The max-pooled features are further processed by $\text{MLP}_2$ with output sizes of $128,128,5+2NH+4NS+NC$ where the output consists of $2$ objectness scores, $3$ center regression values, $2NH$ numbers for heading regression ($NH$ heading bins) and $4NS$ numbers for box size regression ($NS$ box anchors) and $NC$ numbers for semantic classification.

\smallskip\noindent\textbf{Training the network.} We train the entire network end-to-end and from scratch with an Adam optimizer, batch size 8 and an initial learning rate of $0.001$. The learning rate is decreased by $10\times$ after 80 epochs and then decreased by another $10\times$ after 120 epochs. Training the model to convergence on one Volta Quadro GP100 GPU takes around 10 hours on SUN RGB-D and less than 4 hours on ScanNetV2.

\smallskip\noindent\textbf{Inference.} Our \votenet{} is able to take point clouds of the entire scenes and generate proposals in one forward pass. The proposals are post-processed by a 3D NMS module with an IoU threshold of $0.25$. The evaluation follows the same protocol as in~\cite{song2016deep} using mean average precision.

\section{Experiments}
In this section, we firstly compare our Hough voting based detector with previous state-of-the-art methods on two large 3D indoor object detection benchmarks (Sec.~\ref{sec:exp:sota}). We then provide analysis experiments to understand the importance of voting, the effects of different vote aggregation approaches and show our method's advantages in its compactness and efficiency (Sec.~\ref{sec:exp:analysis}). Finally we show qualitative results of our detector (Sec.~\ref{sec:qualitative}). More analysis and visualizations are provided in the appendix.

% To further establish the importance of voting, we provide a thorough ablation study of the different ingredient of our pipeline. To this end, we construct a direct-proposal network that generates detection proposals directly from scene points, i.e. without voting. A conclusion from this study is that voting, when aggregated properly, adds a significant boost to overall performance. In particular, we show that \votenet{} excels in classes where the amodal bounding-box center tends to be far from the object surface (e.g. tables, bathtubs, etc.).

% \begin{itemize}
%     \item Comparison with previous methods on 3D object detection and 3D instance segmentation.
%     \item Key ablation studies to validate our design: voting or not (different objectness losses). to learn vote pooling or not. vote context. vote weighting/masking.
%     \item Visualizations. Intuitive examples to show why voting could help.
%     \item Other experiments: effects of data augmentation. one-stage v.s. two-stage detector (with RoI and box refinement). one v.s. multiple votes. adding rgb to input point channels. 
%     \item Discussion on more applications of \votenet/deep hough voting.
% \end{itemize}

% \rqi{Add a section on evaluation of 3d object proposal: recall of the system.}

\begin{table*}[t!]
\small
\setlength{\tabcolsep}{4.8pt}
\begin{center}
% The model we used is in eval_0310_uptodate_l1_fps_on_seeds2_clsnms_v1data_run4/log_eval.txt evaluated with use_v1, batch_size 1, 3d NMS iou 0.25 and 256 targets.
\begin{tabular}{l|c|x{25}x{25}x{25}x{25}x{25}x{25}x{25}x{25}x{25}x{25}|c}
\toprule
          & Input & bathtub & bed & bookshelf & chair & desk & dresser & nightstand & sofa & table & toilet & mAP \\ \midrule
DSS~\cite{song2016deep} & Geo + RGB & 44.2 & 78.8 & 11.9 & 61.2 & 20.5 & 6.4 & 15.4 & 53.5 & 50.3 & 78.9 & 42.1    \\
COG~\cite{ren2016three} & Geo + RGB & 58.3 & 63.7 & 31.8 & 62.2 & \textbf{45.2} & 15.5 & 27.4 & 51.0 & \textbf{51.3} & 70.1 & 47.6 \\
2D-driven~\cite{lahoud20172d} & Geo + RGB & 43.5 & 64.5 & 31.4 & 48.3 & 27.9 & 25.9 & 41.9 & 50.4 & 37.0 & 80.4 & 45.1  \\
F-PointNet~\cite{qi2018frustum} & Geo + RGB & 43.3 & 81.1 & \textbf{33.3} & 64.2 & 24.7 & \textbf{32.0} & 58.1 & 61.1 & 51.1 & \textbf{90.9} & 54.0 \\ \midrule
\votenet~(ours) & \textbf{Geo only} & \textbf{74.4} & \textbf{83.0} & 28.8 & \textbf{75.3} & 22.0 & 29.8 & \textbf{62.2} & \textbf{64.0} & 47.3 & 90.1 & \textbf{57.7} \\
%Ours (two-stage) (v2) & $\textbf{75.2}$ & $\textbf{83.0}$ & $29.5$ & $\textbf{75.8}$ & $18.1$ & $29.5$ & $\textbf{59.2}$ & $\textbf{66.1}$ & $49.0$ & $89.2$ & N.A. & \textbf{57.5} \\
\bottomrule
\end{tabular}
\end{center}
\caption{\textbf{3D object detection results on SUN RGB-D val set.} Evaluation metric is average precision with 3D IoU threshold 0.25 as proposed by~\cite{song2015sun}. Note that both COG~\cite{ren2016three} and 2D-driven~\cite{lahoud20172d} use room layout context to boost performance. To have fair comparison with previous methods, the evaluation is on the SUN RGB-D V1 data.
% \rqi{Explain why we are not doing well on bookshelf, desk, dresser and table.}
% Compared with previous state-of-the-arts our method is 6.4\% to 11.9\% better in mAP as well as one to three orders of magnitude faster.
}
\label{tab:sunrgbd}
\end{table*}

\begin{table}[t!]
\small
\setlength{\tabcolsep}{3.2pt}
\begin{center}
\begin{tabular}{l |c| c c}
%{ l@{\hskip 0.01\textwidth}c@{\hskip 0.01\textwidth}c@{\hskip 0.01\textwidth}c@{\hskip 0.01\textwidth}c@{\hskip 0.01\textwidth}c@{\hskip 0.01\textwidth}c@{\hskip 0.01\textwidth}c  }
    \toprule
    & Input & mAP@0.25 & mAP@0.5 \\ \hline    
    DSS~\cite{song2016deep,hou20183d} & Geo + RGB & 15.2 & 6.8  \\
    MRCNN 2D-3D~\cite{he2017mask,hou20183d} & Geo + RGB & 17.3 & 10.5 \\
    F-PointNet~\cite{qi2018frustum,hou20183d} & Geo + RGB & 19.8 & 10.8 \\
    GSPN~\cite{yi2018gspn} & Geo + RGB & 30.6 & 17.7 \\ \midrule
    3D-SIS \cite{hou20183d} & Geo + 1 view & 35.1 & 18.7 \\ 
    3D-SIS \cite{hou20183d} & Geo + 3 views & 36.6 & 19.0 \\
    3D-SIS \cite{hou20183d} & Geo + 5 views & 40.2 & 22.5 \\ \midrule
    3D-SIS \cite{hou20183d} & Geo only & 25.4 & 14.6 \\
    % {\bf Ours} & Geo only &$\textbf{46.75} \pm \textbf{0.32}$ & $\textbf{24.65} \pm \textbf{0.58}$ \\ \bottomrule 
    \votenet~(ours) & Geo only & \textbf{58.6} & \textbf{33.5} \\ \bottomrule 
\end{tabular}
\end{center}
%   \vspace{1cm}
\caption{\small \textbf{3D object detection results on ScanNetV2 val set.} DSS and F-PointNet results are from~\cite{hou20183d}. Mask R-CNN 2D-3D results are from~\cite{yi2018gspn}. GSPN and 3D-SIS results are up-to-date numbers provided by the original authors.
}
\label{tab:scannet}
\end{table}

\subsection{Comparing with State-of-the-art Methods}
\label{sec:exp:sota}
% Evaluated on SUN RGB-D \cite{song2015sun} and ScanNetV2 \cite{dai2017scannet}, our method significantly outperforms all previous methods while using purely geometric information. After a short description of the datasets and competing methods, we discuss these results in detail.

% Importantly, we use purely geometric information, suggesting that our algorithm has learned to extract strong semantic and geometric cues directly from 3D. 
% \rqi{Mention that SUN RGB-D are partial scans (single-frame) while ScanNet has scans from multiple RGB-D views so more complete. Our algorithms work well for both cases with the same architecture and the same set of hyperparameters.}

\noindent\textbf{Dataset.}
SUN RGB-D \cite{song2015sun} is a single-view RGB-D dataset for 3D scene understanding. It consists of ${\sim}$5K RGB-D training images %($5,285$ in train set and $5,050$ in val set) 
%annotated with semantic segmentation on images and 
annotated with amodal oriented 3D bounding boxes for $37$ object categories. To feed the data to our network, we firstly convert the depth images to point clouds using the provided camera parameters. We follow a standard evaluation protocol and report performance on the $10$ most common categories.
% As prescribed by previous works, our evaluation is done on the $10$ most common categories with the metric of mean average precision (mAP) using a 3D IoU threshold of $0.25$.

ScanNetV2 \cite{dai2017scannet} is a richly annotated dataset of 3D reconstructed meshes of indoor scenes. It contains ${\sim}$1.2K training examples collected from hundreds of different rooms, and annotated with semantic and instance segmentation for $18$ object categories. Compared to partial scans in SUN RGB-D, scenes in ScanNetV2 are more complete and cover larger areas with more objects on average.
% $1,513$ reconstructed meshes processed from RGB-D videos collected from about seven hundred independent environments. 
% The dataset is annotated with semantic and instance segmentation for $18$ object categories. The public data are split into train and validation sets of size $1.2$k and $312$, respectively. 
We sample vertices from the reconstructed meshes as our input point clouds.  Since ScanNetV2 does not provide amodal or oriented bounding box annotation, we aim to predict axis-aligned bounding boxes instead, as in~\cite{hou20183d}.

\smallskip
\noindent\textbf{Methods in comparison.} We compare with a wide range of prior art methods. Deep sliding shapes (DSS)~\cite{song2016deep} and 3D-SIS~\cite{hou20183d} are both 3D CNN based detectors that combine geometry and RGB cues in object proposal and classification, based on the Faster R-CNN~\cite{ren2015faster} pipeline. Compared with DSS, 3D-SIS introduces a more sophisticated sensor fusion scheme (back-projecting RGB features to 3D voxels) and therefore is able to use multiple RGB views to improve performance. 2D-driven~\cite{lahoud20172d} and F-PointNet~\cite{qi2018frustum} are 2D-based 3D detectors that rely on object detection in 2D images to reduce the 3D detection search space. Cloud of gradients~\cite{ren2016three} is a sliding window based detector using a newly designed 3D HoG-like feature. MRCNN 2D-3D is a na\"ive baseline that directly projects Mask-RCNN~\cite{he2017mask} instance segmentation results into 3D to get a bounding box estimation. GSPN~\cite{yi2018gspn} is a recent instance segmentation method using a generative model to propose object instances, which is also based on a PointNet++ backbone.

\smallskip
\noindent\textbf{Results}
are summarized in Table \ref{tab:sunrgbd} and \ref{tab:scannet}. \votenet{} outperforms all previous methods by at least \textbf{3.7} and \textbf{18.4} mAP increase in SUN RGB-D and ScanNet respectively. Notably, we achieve such improvements when we \emph{use geometric input (point clouds) only} while they used both geometry and RGB images. Table~\ref{tab:sunrgbd} shows that in the category ``chair'' with the most training samples, our method improves upon previous state of the art by more than \textbf{11 AP}.
% , while only falling behind by more than $5\%$ in the desk category, where most other methods struggle.
Table~\ref{tab:scannet} shows that when taking geometric input only, our method outperforms 3D CNN based method 3D-SIS by more than \textbf{33 AP}.
A per-category evaluation for ScanNet is provided in the appendix. Importantly, the same set of network hyper-parameters was used in both datasets. 

% \rqi{Add run time comparison, which we are at advantage.}

\subsection{Analysis Experiments}
\label{sec:exp:analysis}
% We now turn to investigate voting-specific design choices in our network. %: whether or not to vote, vote pooling, and vote sampling for generating proposals. 
% First, we validate the necessity and effectiveness of voting and then compare different aggregation techniques.

% \subsubsection{Voting or Not?}
% \smallskip\noindent\textbf{To Vote or Not To Vote?}
\paragraph{To Vote or Not To Vote?}

% \begin{figure}
%     \centering
%     \includegraphics[width=\linewidth]{./fig/vote_or_not}
%     \caption{From left to right: propose object locations from rasterized pixels or voxels, the strategy taken by 3D CNN detectors; propose objects from boundary points, the approach of \boxnet; propose objects from aggregated votes, which is our \votenet's appraoch.}
%     \label{fig:my_label}
% \end{figure}

% \begin{table}[t!]
%     \small
%     \begin{center}
%     \begin{tabular}{c|c|cc}
%     \toprule
%         Method & 3D  & \multicolumn{2}{c}{mAP@0.25} \\
%         & representation & SUN RGB-D & ScanNet \\ \midrule
%         DSS~\cite{song2016deep} & Volumetric & 42.1 & 15.2 \\ 
%         3D-SIS~\cite{hou20183d} & Volumetric &  - & 25.4 \\ \midrule
%         \boxnet~(ours) & Point clouds & 53.0 & 39.6 \\ \midrule
%         \votenet~(ours) & Point clouds & \textbf{57.7} & \textbf{46.8} \\
%         \bottomrule
%     \end{tabular}
%     \end{center}
%     \caption{\textbf{Comparing methods with different proposal strategies.} 3D CNNs propose from dense voxels; \votenet~proposes from vote clusters; \boxnet~(our baseline model without voting) proposes boxes from seed points on object surfaces.}
%     \label{tab:vote_or_not}
% \end{table}

\begin{table}[t!]
    \small
    \begin{center}
    \begin{tabular}{c|cc}
    \toprule
        Method & \multicolumn{2}{c}{mAP@0.25} \\
        & SUN RGB-D & ScanNet \\ \midrule
        % \boxnet~(ours) & 53.0 & 39.6 \\ \midrule
         \boxnet~(ours) & 53.0 & 45.4 \\ \midrule
        \votenet~(ours) & \textbf{57.7} & \textbf{58.6} \\
        \bottomrule
    \end{tabular}
    \end{center}
    \caption{\textbf{Comparing \votenet{} with a no-vote baseline.} Metric is 3D object detection mAP. \votenet~estimate object bounding boxes from vote clusters. \boxnet~ proposes boxes directly from seed points on object surfaces without voting.}
    \label{tab:vote_or_not}
\end{table}

A straightforward baseline to \votenet{} is a network that directly proposes boxes from sampled scene points. Such a baseline -- which we refer to as \emph{\boxnet{}} -- is essential to distill the improvement due to voting. The \boxnet~has the same backbone as the \votenet{} but instead of voting, it directly generates boxes from the seed points (more details in appendix). Table~\ref{tab:vote_or_not} shows voting boosts the performance by a significant margin of ${\sim}$5 mAP on SUN RGB-D and $>$13 mAP on ScanNet.

%As seen in Table~\ref{tab:vote_or_not}, the \boxnet{} already outperforms all previous geometry-based methods by a large margin, and is on-par with the best performing RGB-D based ones. Yet, we

\begin{figure}[t!]
    \centering
    \begin{overpic}
    [trim=0cm 0cm 0cm 0cm,clip,width=0.8\linewidth]{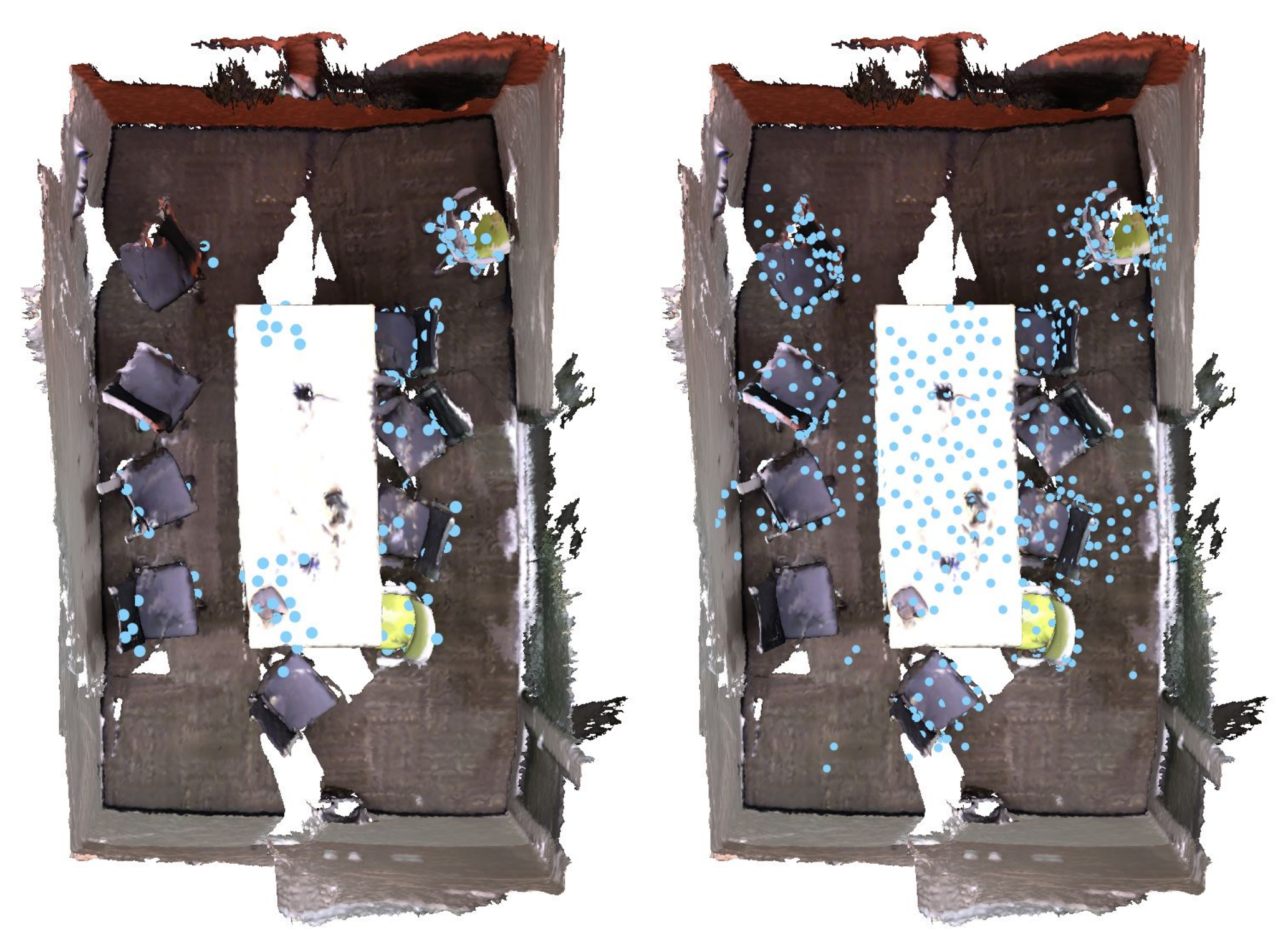}
    \put(8,73){\small \boxnet{} (no voting)}
    \put(65,73){\small \votenet{}}
    \end{overpic}
    \caption{\textbf{Voting helps increase detection contexts.} Seed points that generate good boxes (\boxnet), or good votes (\votenet) which in turn generate good boxes, are overlaid (in blue) on top of a representative ScanNet scene. As the voting step effectively increases context, \votenet{} demonstrates a much denser cover of the scene, therefore increasing the likelihood of accurate detection.}
    \label{fig:voting_vs_novote}
\end{figure}

In what ways, then, does voting help? We argue that since in sparse 3D point clouds, existing scene points are often far from object centroids, a direct proposal may have lower confidence and inaccurate amodal boxes. Voting, instead, brings closer together these lower confidence points and allows to reinforce their hypothesis though aggregation. We demonstrate this phenomenon in Fig.~\ref{fig:voting_vs_novote} on a typical ScanNetV2 scene. We overlay the scene with only those seed points which, if sampled, would generate an accurate proposal. As can be seen, \votenet{} (right) offers a much broader coverage of ``good'' seed points compared to \boxnet{} (left), showing its robustness brought by voting.

\begin{figure}[t!]
    \centering
    \includegraphics[width=0.85\linewidth]{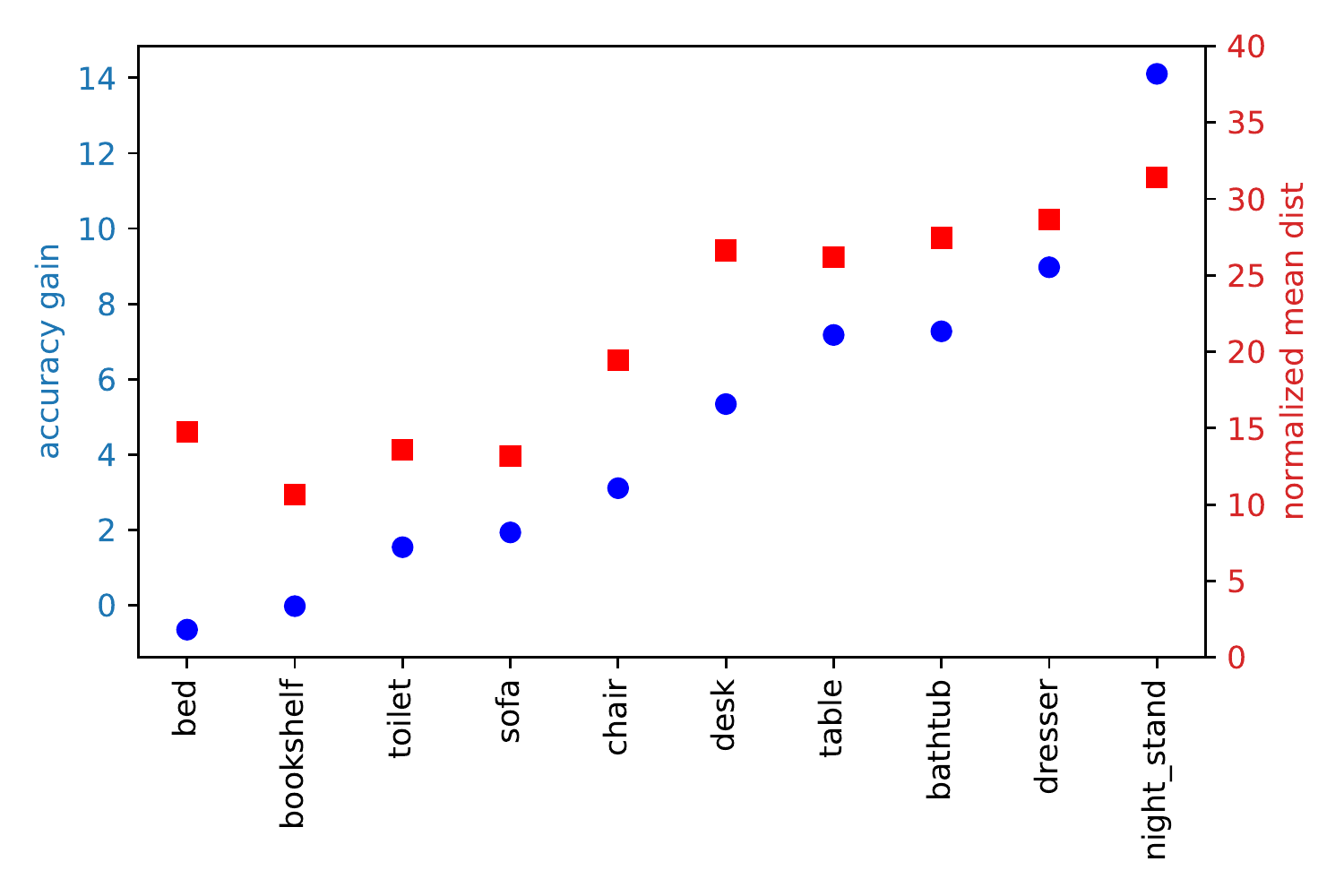}
    \caption{\textbf{Voting helps more in cases where object points are far from object centers}. We show for each category: voting accuracy gain (in blue dots) of \votenet{} w.r.t our direct proposal baseline \boxnet{}; and (in red squares) average object-center distance, normalized by the mean class size.}
    \label{fig:perclass}
\end{figure}

We proceed with a second analysis in Fig.~\ref{fig:perclass} showing on the same plot (in separate scales), for each SUN RGB-D category: (in blue dots) gains in mAP between \votenet{} and \boxnet{}, and (in red squares) closest distances between object points (on their surfaces) and their amodal box centers, averaged per category and normalized by the mean class size (a large distance means the object center is usually far from its surface). Sorting the categories according to the former, we see a strong correlation. Namely, when object points tend to be further from the amodal box center, voting helps much more.

% [TBD] Message 1: Voting is better. Point cloud based methods are more effective (even our baseline is on par or better compared with previous state of the arts).
% Message 2: Voting is better in two ways: one is that it makes a fuller use of information. See~\ref{fig:voting_vs_novote}; two is that for certain classes like night stand, dresser and table, existing points are far from object centers, so voting based method has a better chance in localizing them more accurately. See~\ref{fig:perclass}.

% \begin{table}[]
%     \small
%     \centering
%     \begin{tabular}{c|c|cc}
%     \toprule
%         Method & Objectness & \multicolumn{2}{c}{mAP@0.25} \\
%         & & SUN RGB-D & ScanNet \\ \midrule
%         DSS~\cite{song2016deep} & IoU & $42.1$ & $15.2$ \\ 
%         3D-SIS~\cite{hou20183d} & IoU & - & $25.4$ \\ \midrule
%         \multirow{3}{*}{BoxNet (ours)} & Box XYZ & & \\
%         & Seed XYZ & & \\
%         & Seed class & & \\ \midrule
%         \multirow{3}{*}{\votenet (ours)} & Box XYZ & & \\
%         & Vote XYZ & & \\
%         & Vote class & & \\
%         \bottomrule
%     \end{tabular}
%     \caption{Comparing \votenet with BoxNet (our baseline model without voting) and 3D CNN detectors.}
%     \label{tab:vote_or_not}
% \end{table}

% \smallskip\noindent\textbf{Effect of Vote Aggregation}
\paragraph{Effect of Vote Aggregation}
\label{sec:aggregation_variants}
Aggregation of votes is an important component in \votenet{} as it allows communication between votes. Hence, it is useful to analyze how different aggregation schemes influence performance. 

% Although both \boxnet{} and our voting-based method regress an offset from boundary points to object centers, the major difference between them is that regressed votes are aggregated in the \votenet while regressed boxes are not combined in the \boxnet. 
% In this section, we show how vote aggregation matters for the 3D object detection performance and compare our aggregation with a few alternatives.

%With radius equal to $0$, we do not aggregate votes for proposals -- each vote independently proposes an object bounding box, which is equivalent to the~\boxnet. As we increase the attention region by increasing the radius in vote aggregation, we see the~\votenet performance increases and peaks at around $0.2$ radius. 

\begin{figure}[t!]
\begin{minipage}[c]{0.52\linewidth}
\begin{center}
\includegraphics[width=\linewidth]{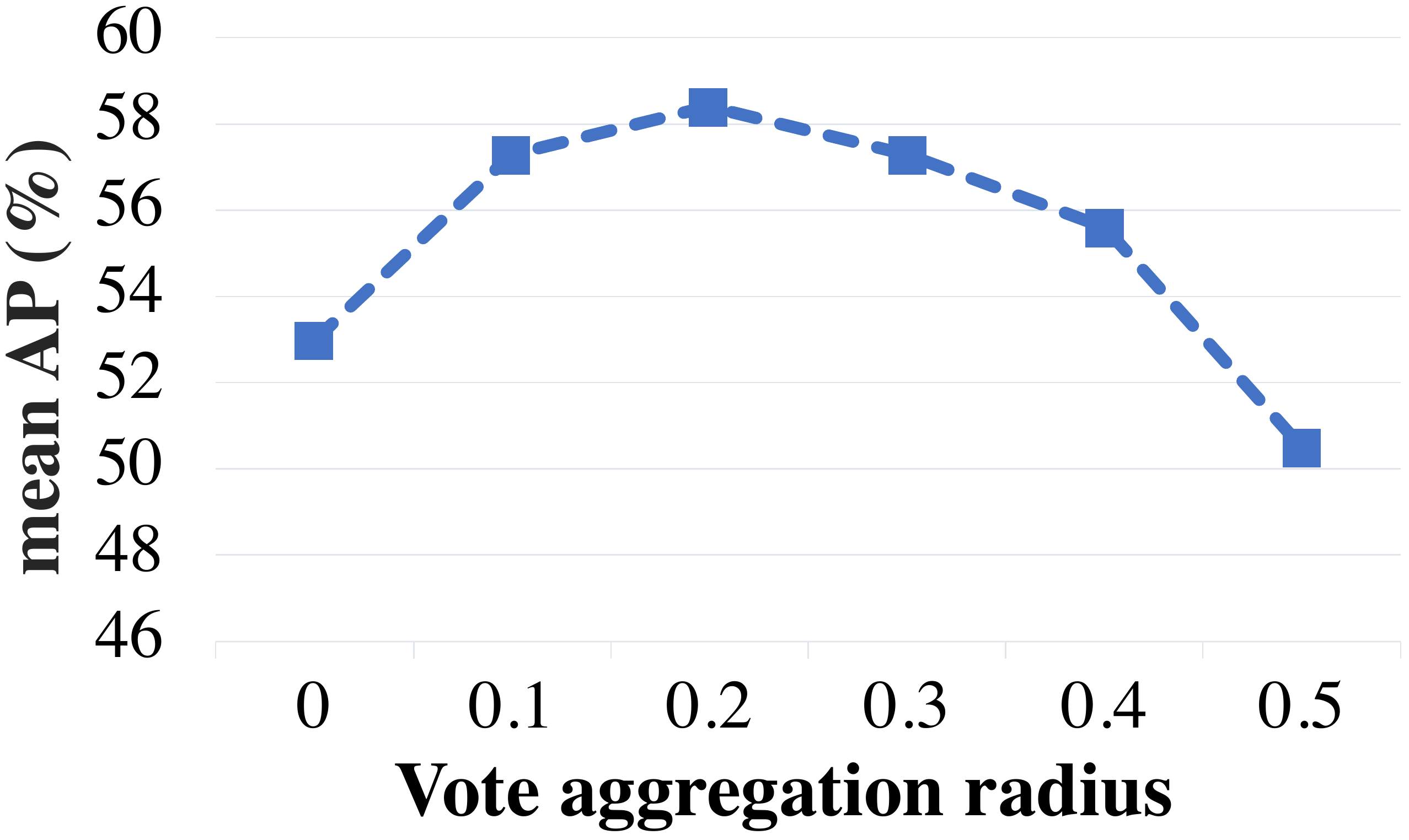}
\end{center}
\end{minipage}%
\begin{minipage}[c]{0.48\linewidth}
\begin{center}
{
\fontsize{8pt}{1em}\selectfont
\begin{tabular}{c|c}
\toprule
  Aggregation method & mAP \\
\midrule
Feature avg. & 47.2 \\ 
Feature max & 47.8 \\
Feature RBF avg. & 49.0 \\
\midrule
Pointnet (avg.) & 56.5 \\
Pointnet (max) & 57.7 \\
\bottomrule
\end{tabular}
}
\end{center}
\end{minipage}
\vspace{0.1in}
    \caption{\textbf{Vote aggregation analysis.} \emph{Left:} mAP@0.25 on SUN RGB-D for varying aggregation radii when aggregating via Pointnet (max). \emph{Right:} Comparisons of different aggregation methods (radius = $0.3$ for all methods). Using a learned vote aggregation is far more effective than manually pooling the features in a local neighborhood.}
    \label{fig:vote_pooling}
\end{figure}

In Fig.~\ref{fig:vote_pooling} (right), we show that vote aggregation with a learned Pointnet and max pooling achieves far better results than manually aggregating the vote features in the local regions due to the existence of clutter votes (i.e. votes from non-object seeds). We test 3 types of those aggregations (first three rows): max, average, and RBF weighting (based on vote distance to the cluster center). In contrast to aggregation with Pointnet (Eq.~\ref{eq:vote_aggregation}), the vote features are directly pooled, e.g. for avg. pooling: $p = \text{MLP}_2 \left\{ \mbox{AVG}\{h_i\} \right\}$).

In Fig.~\ref{fig:vote_pooling} (left), we show how vote aggregation radius affects detection (tested with Pointent using max pooling). As the aggregation radius increases, \votenet{} improves until it peaks at around $0.2$ radius. Attending to a larger region though introduces more clutter votes thus contaminating the good votes and results in decreased performance.

\paragraph{Model Size and Speed}
Our proposed model is very efficient since it leverages sparsity in point clouds and avoids search in empty space. Compared to previous best methods (Table~\ref{tab:size_speed}), our model is more than $4 \times$ smaller than F-PointNet (the prior art on SUN RGB-D) in size and more than $20 \times$ times faster than 3D-SIS (the prior art on ScanNetV2) in speed. Note that the ScanNetV2 processing time by 3D-SIS is computed as averaged time in offline batch mode while ours is measured with sequential processing which can be realized in online applications.

\begin{table}[b!]
\small
    \begin{center}
    \begin{tabular}{l|c|c|c}
    \toprule
        Method & Model size & SUN RGB-D & ScanNetV2 \\ \midrule
        F-PointNet~\cite{qi2018frustum} & $47.0$MB & $0.09s$ & -\\
        3D-SIS~\cite{hou20183d} & $19.7$MB & - & $2.85$s \\ \midrule
        \votenet~(ours) & 11.2MB & $0.10s$ & $0.14$s\\ \bottomrule
    \end{tabular}
    \end{center}
    \caption{\textbf{Model size and processing time (per frame or scan).} Our method is more than $4 \times$ more compact in model size than~\cite{qi2018frustum} and more than $20 \times$ faster than~\cite{hou20183d}.}
    \label{tab:size_speed}
\end{table}

\begin{figure*}[t!]
    \centering
    \begin{overpic}
    [trim=0cm 0cm 0cm 0cm,clip,width=0.86\linewidth]{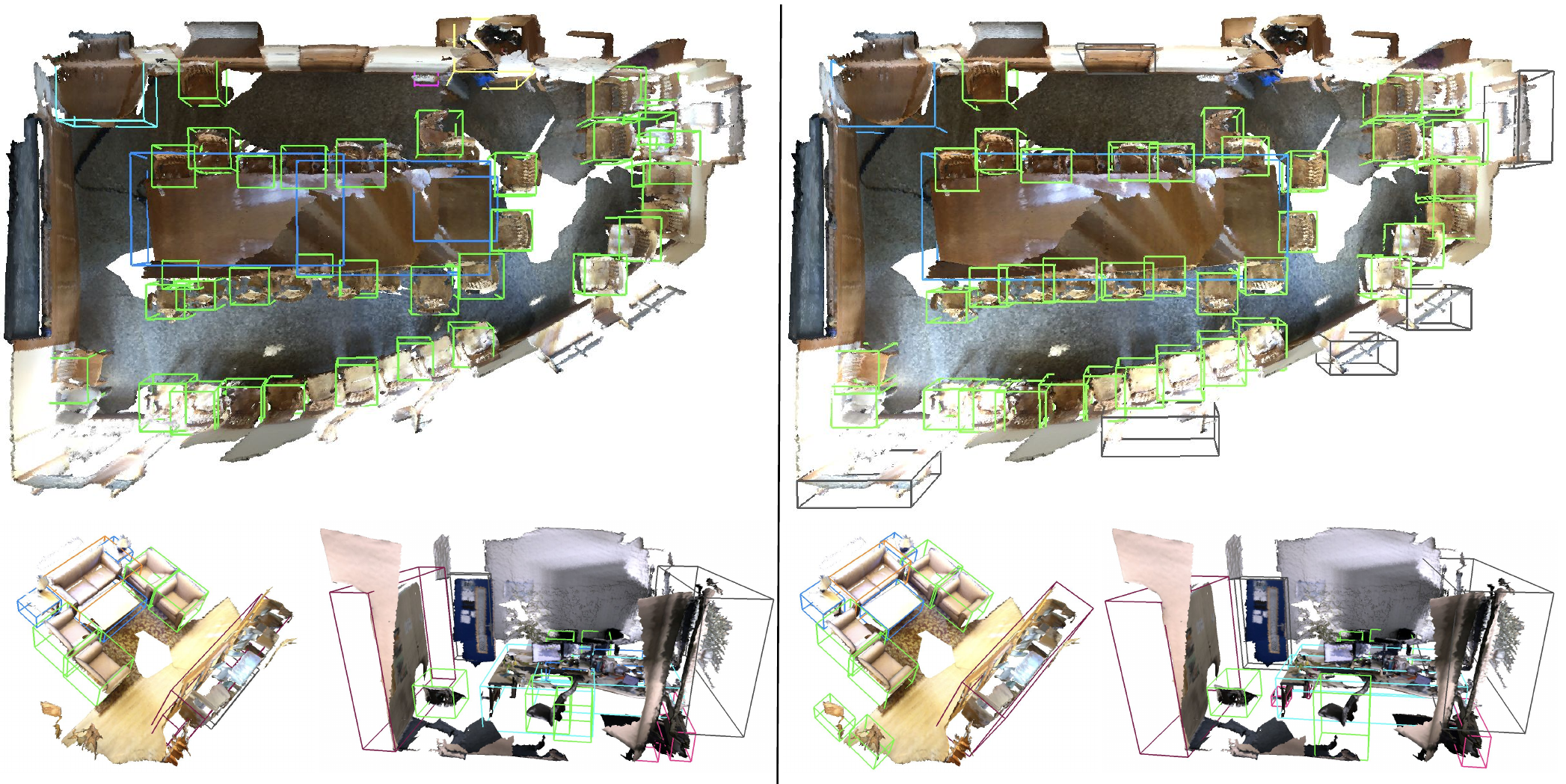}
    \put(15,50){\small \votenet{} prediction}
    \put(70,50){\small Ground truth}
    \end{overpic}
    \caption{\textbf{Qualitative results of 3D object detection in ScanNetV2.} Left: our \votenet{}, Right: ground-truth. See Section \ref{sec:qualitative} for details.}
    \label{fig:qualitative_results}
\end{figure*}

\begin{figure*}[t!]
    \centering
    \includegraphics[width=0.86\linewidth]{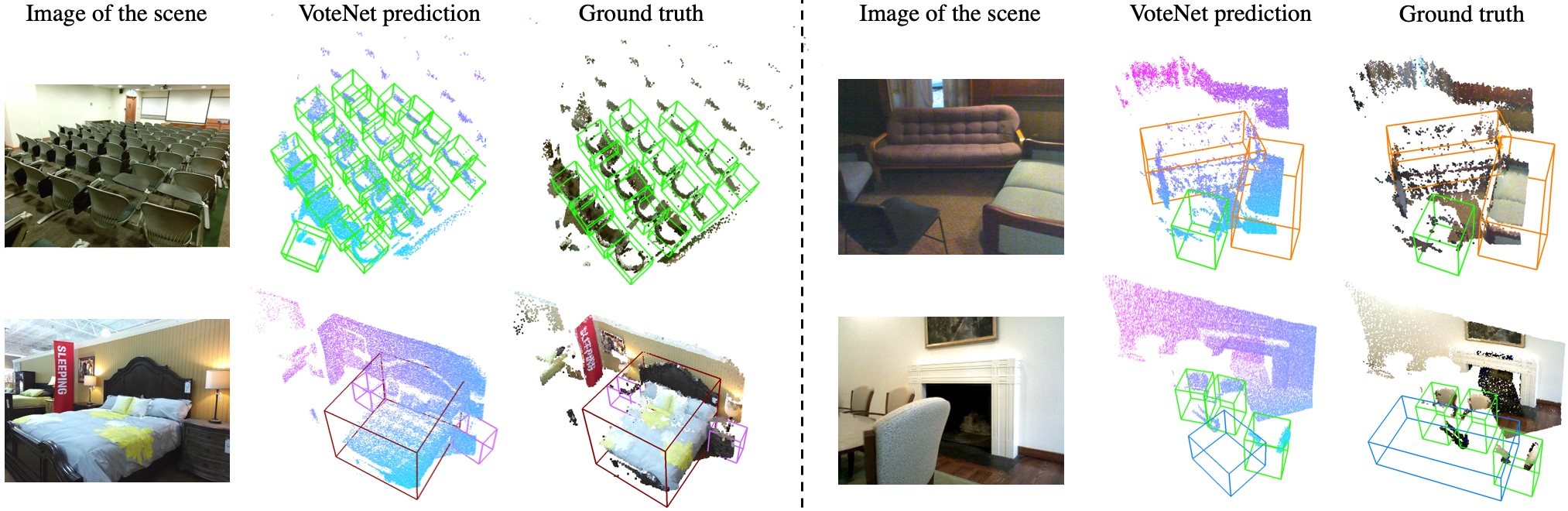}
    \caption{\textbf{Qualitative results on SUN RGB-D.} Both left and right panels show (from left to right): an image of the scene (not used by our network), 3D object detection by \votenet{}, and ground-truth annotations. See Section \ref{sec:qualitative} for details.}
    \label{fig:qualitative_results2}
\end{figure*}

% % \subsubsection{Effects of Vote Cluster Sampling}
% % \begin{table}[]
% %     \small
% %     \centering
% %     \begin{tabular}{c|c|c}
% %     \toprule
% %         Vote cluster sampling & Vote weighting & mAP \\ \midrule
% %         Farthest point sampling & None & $56.7$\\
% %         Random sampling & None & $57.2$\\
% %         FPS on seeds & None & $57.2$\\
% %         FPS on seeds & Yes & $57.7$\\
% %         \bottomrule
% %     \end{tabular}
% %     \caption{Effects of vote cluster sampling and vote weighting. \rqi{Currently the ``score weighted sampling'' is just seed FPS sampling. We need a better sampling strategy or a more convincing analysis here.}}
% %     \label{tab:seed_masking}
% % \end{table}

\subsection{Qualitative Results and Discussion}
\label{sec:qualitative}
Fig.~\ref{fig:qualitative_results} and Fig.~\ref{fig:qualitative_results2} show several representative examples of \votenet{} detection results on ScanNet and SUN RGB-D scenes, respectively. As can be seen, the scenes are quite diverse and pose multiple challenges including clutter, partiality, scanning artifacts, etc. Despite these challenges, our network demonstrates quite robust results. See for example in Fig.~\ref{fig:qualitative_results}, how the vast majority of chairs were correctly detected in the top scene. Our method was able to nicely distinguish between the attached sofa-chairs and the sofa in the bottom left scene; and predicted the complete bounding box of the much fragmented and cluttered desk at the bottom right scene.

There are still limitations in our method though. Common failure cases include misses on very thin objects like doors, windows and pictures denoted in black bounding boxes in the top scene (Fig.~\ref{fig:qualitative_results}). As we do not make use of RGB information, detecting these categories is almost impossible. Fig.~\ref{fig:qualitative_results2} on SUN RGB-D also reveals the strengths of our method in partial scans with single-view depth images. For example, it detected more chairs in the top-left scene than were provided by the ground-truth. In the top-right scene we can see how \votenet{} can nicely hallucinate the amodal bounding box despite seeing only part of the sofa. A less successful amodal prediction is shown in the bottom right scene where an extremely partial observation of a very large table is given.

\section{Conclusion}
% the purpose: we aimed to squeeze as much as possible from geometry. 
% When lifting the problem of object detection from 2D to 3D, an inherent challenge emerges: object centroids can be distant from its surface points and are often vacant. To address this challenge while preserving computational efficiency
In this work we have introduced \votenet{}: a simple, yet powerful 3D object detection model inspired by Hough voting. The network learns to vote to object centroids directly from point clouds and learns to aggregate votes through their features and local geometry to generate high-quality object proposals. Using only 3D point clouds, the model showed significant improvements over previous methods that utilize both depth and colored images.
% Importantly, the voting layer is general and can be assumed with other backbone architectures.
% Finally, voting can be interpreted as a form of connecting scene points to be processed together, by embedding learned features into $3$ dimensions.
% This calls to further investigate this technique from a Graph-CNN viewpoint.
% In future work we intend to explore these directions, as well as means to incorporate RGB information. We are also interested in utilizing our detector in downstream application such as semantic and instance segmentation.  We also believe that the synergy of Hough voting and deep learning can be applicable to more applications such as 6D pose estimation, template based detection etc. and expect to see more future research along this line.

In future work we intend to explore how to incorporate RGB images into our detection framework and to utilize our detector in downstream application such as 3D instance segmentation. We believe that the synergy of Hough voting and deep learning can be generalized to more applications such as 6D pose estimation, template based detection etc. and expect to see more future research along this line.

\paragraph{Ackownledgements.}
This work was supported in part by ONR MURI grant N00014-13-1-0341, NSF grant IIS-1763268 and a Vannevar Bush Faculty Fellowship. We thank Daniel Huber, Justin Johnson, Georgia Gkioxari and Jitendra Malik for valuable discussions and feedback.

%%%%%%%%% Acknowledgements
% Daniel Huber, Justin Johnson, Jitendra Malik, ...

%%%%%%%%% REFERENCES
% \clearpage

{\small
\bibliographystyle{ieee_fullname}
\bibliography{pcl}

\begin{thebibliography}{10}\itemsep=-1pt

\bibitem{atzmon2018point}
Matan Atzmon, Haggai Maron, and Yaron Lipman.
\newblock Point convolutional neural networks by extension operators.
\newblock {\em arXiv preprint arXiv:1803.10091}, 2018.

\bibitem{ballard1981generalizing}
Dana~H Ballard.
\newblock Generalizing the hough transform to detect arbitrary shapes.
\newblock {\em Pattern recognition}, 13(2):111--122, 1981.

\bibitem{borrmann20113d}
Dorit Borrmann, Jan Elseberg, Kai Lingemann, and Andreas N{\"u}chter.
\newblock The 3d hough transform for plane detection in point clouds: A review
  and a new accumulator design.
\newblock {\em 3D Research}, 2(2):3, 2011.

\bibitem{cvpr17chen}
Xiaozhi Chen, Huimin Ma, Ji Wan, Bo Li, and Tian Xia.
\newblock Multi-view 3d object detection network for autonomous driving.
\newblock In {\em IEEE CVPR}, 2017.

\bibitem{dai2017scannet}
Angela Dai, Angel~X Chang, Manolis Savva, Maciej Halber, Thomas Funkhouser, and
  Matthias Nie{\ss}ner.
\newblock Scannet: Richly-annotated 3d reconstructions of indoor scenes.
\newblock In {\em Proceedings of the IEEE Conference on Computer Vision and
  Pattern Recognition}, pages 5828--5839, 2017.

\bibitem{fan2017point}
Haoqiang Fan, Hao Su, and Leonidas~J Guibas.
\newblock A point set generation network for 3d object reconstruction from a
  single image.
\newblock In {\em Proceedings of the IEEE conference on computer vision and
  pattern recognition}, pages 605--613, 2017.

\bibitem{gall2013class}
Juergen Gall and Victor Lempitsky.
\newblock Class-specific hough forests for object detection.
\newblock In {\em Decision forests for computer vision and medical image
  analysis}, pages 143--157. Springer, 2013.

\bibitem{gall2011hough}
Juergen Gall, Angela Yao, Nima Razavi, Luc Van~Gool, and Victor Lempitsky.
\newblock Hough forests for object detection, tracking, and action recognition.
\newblock {\em IEEE transactions on pattern analysis and machine intelligence},
  33(11):2188--2202, 2011.

\bibitem{graham20183d}
Benjamin Graham, Martin Engelcke, and Laurens van~der Maaten.
\newblock 3d semantic segmentation with submanifold sparse convolutional
  networks.
\newblock In {\em Proceedings of the IEEE Conference on Computer Vision and
  Pattern Recognition}, pages 9224--9232, 2018.

\bibitem{guerrero2018pcpnet}
Paul Guerrero, Yanir Kleiman, Maks Ovsjanikov, and Niloy~J Mitra.
\newblock Pcpnet learning local shape properties from raw point clouds.
\newblock In {\em Computer Graphics Forum}, volume~37, pages 75--85. Wiley
  Online Library, 2018.

\bibitem{he2017mask}
Kaiming He, Georgia Gkioxari, Piotr Doll{\'a}r, and Ross Girshick.
\newblock Mask {R-CNN}.
\newblock {\em arXiv preprint arXiv:1703.06870}, 2017.

\bibitem{hou20183d}
Ji Hou, Angela Dai, and Matthias Nie{\ss}ner.
\newblock {3D-SIS}: 3d semantic instance segmentation of rgb-d scans.
\newblock {\em arXiv preprint arXiv:1812.07003}, 2018.

\bibitem{hough1959machine}
Paul~VC Hough.
\newblock Machine analysis of bubble chamber pictures.
\newblock In {\em Conf. Proc.}, volume 590914, pages 554--558, 1959.

\bibitem{huan2017vehicle}
Li Huan, Qin Yujian, and Wang Li.
\newblock Vehicle logo retrieval based on hough transform and deep learning.
\newblock In {\em Proceedings of the IEEE International Conference on Computer
  Vision}, pages 967--973, 2017.

\bibitem{kehl2016deep}
Wadim Kehl, Fausto Milletari, Federico Tombari, Slobodan Ilic, and Nassir
  Navab.
\newblock Deep learning of local rgb-d patches for 3d object detection and 6d
  pose estimation.
\newblock In {\em European Conference on Computer Vision}, pages 205--220.
  Springer, 2016.

\bibitem{kim2013accurate}
Byung-soo Kim, Shili Xu, and Silvio Savarese.
\newblock Accurate localization of 3d objects from rgb-d data using
  segmentation hypotheses.
\newblock In {\em Proceedings of the IEEE Conference on Computer Vision and
  Pattern Recognition}, pages 3182--3189, 2013.

\bibitem{klokov2017escape}
Roman Klokov and Victor Lempitsky.
\newblock Escape from cells: Deep kd-networks for the recognition of 3d point
  cloud models.
\newblock In {\em Proceedings of the IEEE International Conference on Computer
  Vision}, pages 863--872, 2017.

\bibitem{knopp2010orientation}
Jan Knopp, Mukta Prasad, and Luc Van~Gool.
\newblock Orientation invariant 3d object classification using hough transform
  based methods.
\newblock In {\em Proceedings of the ACM workshop on 3D object retrieval},
  pages 15--20. ACM, 2010.

\bibitem{knopp2011scene}
Jan Knopp, Mukta Prasad, and Luc Van~Gool.
\newblock Scene cut: Class-specific object detection and segmentation in 3d
  scenes.
\newblock In {\em 2011 International Conference on 3D Imaging, Modeling,
  Processing, Visualization and Transmission}, pages 180--187. IEEE, 2011.

\bibitem{lahoud20172d}
Jean Lahoud and Bernard Ghanem.
\newblock 2d-driven 3d object detection in rgb-d images.
\newblock In {\em Proceedings of the IEEE Conference on Computer Vision and
  Pattern Recognition}, pages 4622--4630, 2017.

\bibitem{landrieu2018large}
Loic Landrieu and Martin Simonovsky.
\newblock Large-scale point cloud semantic segmentation with superpoint graphs.
\newblock In {\em Proceedings of the IEEE Conference on Computer Vision and
  Pattern Recognition}, pages 4558--4567, 2018.

\bibitem{le2018pointgrid}
Truc Le and Ye Duan.
\newblock Pointgrid: A deep network for 3d shape understanding.
\newblock In {\em Proceedings of the IEEE conference on computer vision and
  pattern recognition}, pages 9204--9214, 2018.

\bibitem{leibe2004combined}
Bastian Leibe, Ales Leonardis, and Bernt Schiele.
\newblock Combined object categorization and segmentation with an implicit
  shape model.
\newblock In {\em Workshop on statistical learning in computer vision, ECCV},
  volume~2, page~7, 2004.

\bibitem{leibe2008robust}
Bastian Leibe, Ale{\v{s}} Leonardis, and Bernt Schiele.
\newblock Robust object detection with interleaved categorization and
  segmentation.
\newblock {\em International journal of computer vision}, 77(1-3):259--289,
  2008.

\bibitem{li2018pointcnn}
Yangyan Li, Rui Bu, Mingchao Sun, Wei Wu, Xinhan Di, and Baoquan Chen.
\newblock Pointcnn: Convolution on x-transformed points.
\newblock In {\em Advances in Neural Information Processing Systems}, pages
  828--838, 2018.

\bibitem{li2015database}
Yangyan Li, Angela Dai, Leonidas Guibas, and Matthias Nie{\ss}ner.
\newblock Database-assisted object retrieval for real-time 3d reconstruction.
\newblock In {\em Computer Graphics Forum}, volume~34. Wiley Online Library,
  2015.

\bibitem{lin2013holistic}
Dahua Lin, Sanja Fidler, and Raquel Urtasun.
\newblock Holistic scene understanding for 3d object detection with rgbd
  cameras.
\newblock In {\em Proceedings of the IEEE International Conference on Computer
  Vision}, pages 1417--1424, 2013.

\bibitem{litany2015asist}
Or Litany, Tal Remez, Daniel Freedman, Lior Shapira, Alex Bronstein, and Ran
  Gal.
\newblock Asist: automatic semantically invariant scene transformation.
\newblock {\em CVIU}, 157:284--299, 2017.

\bibitem{liu2016ssd}
Wei Liu, Dragomir Anguelov, Dumitru Erhan, Christian Szegedy, Scott Reed,
  Cheng-Yang Fu, and Alexander~C Berg.
\newblock Ssd: Single shot multibox detector.
\newblock In {\em European conference on computer vision}, pages 21--37.
  Springer, 2016.

\bibitem{maji2009object}
Subhransu Maji and Jitendra Malik.
\newblock Object detection using a max-margin hough transform.
\newblock 2009.

\bibitem{milletari2017hough}
Fausto Milletari, Seyed-Ahmad Ahmadi, Christine Kroll, Annika Plate, Verena
  Rozanski, Juliana Maiostre, Johannes Levin, Olaf Dietrich, Birgit
  Ertl-Wagner, Kai B{\"o}tzel, et~al.
\newblock Hough-cnn: deep learning for segmentation of deep brain regions in
  mri and ultrasound.
\newblock {\em Computer Vision and Image Understanding}, 164:92--102, 2017.

\bibitem{Indoor2012}
Liangliang Nan, Ke Xie, and Andrei Sharf.
\newblock A search-classify approach for cluttered indoor scene understanding.
\newblock {\em ACM Transactions on Graphics (Proceedings of SIGGRAPH Asia
  2012)}, 31(6), 2012.

\bibitem{novotny2018semi}
David Novotny, Samuel Albanie, Diane Larlus, and Andrea Vedaldi.
\newblock Semi-convolutional operators for instance segmentation.
\newblock In {\em Proceedings of the European Conference on Computer Vision
  (ECCV)}, pages 86--102, 2018.

\bibitem{qi2018frustum}
Charles~R Qi, Wei Liu, Chenxia Wu, Hao Su, and Leonidas~J Guibas.
\newblock Frustum pointnets for 3d object detection from rgb-d data.
\newblock In {\em Proceedings of the IEEE Conference on Computer Vision and
  Pattern Recognition}, pages 918--927, 2018.

\bibitem{qi2017pointnet}
Charles~R Qi, Hao Su, Kaichun Mo, and Leonidas~J Guibas.
\newblock Pointnet: Deep learning on point sets for 3d classification and
  segmentation.
\newblock {\em Proc. Computer Vision and Pattern Recognition (CVPR), IEEE},
  2017.

\bibitem{qi2017pointnetplusplus}
Charles~R Qi, Li Yi, Hao Su, and Leonidas~J Guibas.
\newblock Pointnet++: Deep hierarchical feature learning on point sets in a
  metric space.
\newblock {\em arXiv preprint arXiv:1706.02413}, 2017.

\bibitem{ren2015faster}
Shaoqing Ren, Kaiming He, Ross Girshick, and Jian Sun.
\newblock Faster {R-CNN}: Towards real-time object detection with region
  proposal networks.
\newblock In {\em Advances in neural information processing systems}, pages
  91--99, 2015.

\bibitem{ren2016three}
Zhile Ren and Erik~B Sudderth.
\newblock Three-dimensional object detection and layout prediction using clouds
  of oriented gradients.
\newblock In {\em Proceedings of the IEEE Conference on Computer Vision and
  Pattern Recognition}, pages 1525--1533, 2016.

\bibitem{shi2018pointrcnn}
Shaoshuai Shi, Xiaogang Wang, and Hongsheng Li.
\newblock Pointrcnn: 3d object proposal generation and detection from point
  cloud.
\newblock {\em arXiv preprint arXiv:1812.04244}, 2018.

\bibitem{song2015sun}
Shuran Song, Samuel~P Lichtenberg, and Jianxiong Xiao.
\newblock Sun rgb-d: A rgb-d scene understanding benchmark suite.
\newblock In {\em Proceedings of the IEEE Conference on Computer Vision and
  Pattern Recognition}, pages 567--576, 2015.

\bibitem{song2014sliding}
Shuran Song and Jianxiong Xiao.
\newblock Sliding shapes for 3d object detection in depth images.
\newblock In {\em Computer Vision--ECCV 2014}, pages 634--651. Springer, 2014.

\bibitem{song2016deep}
Shuran Song and Jianxiong Xiao.
\newblock Deep sliding shapes for amodal 3d object detection in rgb-d images.
\newblock In {\em Proceedings of the IEEE Conference on Computer Vision and
  Pattern Recognition}, pages 808--816, 2016.

\bibitem{su2018splatnet}
Hang Su, Varun Jampani, Deqing Sun, Subhransu Maji, Evangelos Kalogerakis,
  Ming-Hsuan Yang, and Jan Kautz.
\newblock Splatnet: Sparse lattice networks for point cloud processing.
\newblock In {\em Proceedings of the IEEE Conference on Computer Vision and
  Pattern Recognition}, pages 2530--2539, 2018.

\bibitem{sun2010depth}
Min Sun, Gary Bradski, Bing-Xin Xu, and Silvio Savarese.
\newblock Depth-encoded hough voting for joint object detection and shape
  recovery.
\newblock In {\em European Conference on Computer Vision}, pages 658--671.
  Springer, 2010.

\bibitem{tatarchenko2017octree}
Maxim Tatarchenko, Alexey Dosovitskiy, and Thomas Brox.
\newblock Octree generating networks: Efficient convolutional architectures for
  high-resolution 3d outputs.
\newblock {\em arXiv preprint arXiv:1703.09438}, 2017.

\bibitem{tatarchenko2018tangent}
Maxim Tatarchenko, Jaesik Park, Vladlen Koltun, and Qian-Yi Zhou.
\newblock Tangent convolutions for dense prediction in 3d.
\newblock In {\em Proceedings of the IEEE Conference on Computer Vision and
  Pattern Recognition}, pages 3887--3896, 2018.

\bibitem{velizhev2012implicit}
Alexander Velizhev, Roman Shapovalov, and Konrad Schindler.
\newblock Implicit shape models for object detection in 3d point clouds.
\newblock In {\em International Society of Photogrammetry and Remote Sensing
  Congress}, volume~2, 2012.

\bibitem{wang2017cnn}
Peng-Shuai Wang, Yang Liu, Yu-Xiao Guo, Chun-Yu Sun, and Xin Tong.
\newblock O-cnn: Octree-based convolutional neural networks for 3d shape
  analysis.
\newblock {\em ACM Transactions on Graphics (TOG)}, 36(4):72, 2017.

\bibitem{wang2018dynamic}
Yue Wang, Yongbin Sun, Ziwei Liu, Sanjay~E Sarma, Michael~M Bronstein, and
  Justin~M Solomon.
\newblock Dynamic graph cnn for learning on point clouds.
\newblock {\em arXiv preprint arXiv:1801.07829}, 2018.

\bibitem{woodford2014demisting}
Oliver~J Woodford, Minh-Tri Pham, Atsuto Maki, Frank Perbet, and Bj{\"o}rn
  Stenger.
\newblock Demisting the hough transform for 3d shape recognition and
  registration.
\newblock {\em International Journal of Computer Vision}, 106(3):332--341,
  2014.

\bibitem{xie2018attentional}
Saining Xie, Sainan Liu, Zeyu Chen, and Zhuowen Tu.
\newblock Attentional shapecontextnet for point cloud recognition.
\newblock In {\em Proceedings of the IEEE Conference on Computer Vision and
  Pattern Recognition}, pages 4606--4615, 2018.

\bibitem{xu2018spidercnn}
Yifan Xu, Tianqi Fan, Mingye Xu, Long Zeng, and Yu Qiao.
\newblock Spidercnn: Deep learning on point sets with parameterized
  convolutional filters.
\newblock In {\em Proceedings of the European Conference on Computer Vision
  (ECCV)}, pages 87--102, 2018.

\bibitem{yang2018foldingnet}
Yaoqing Yang, Chen Feng, Yiru Shen, and Dong Tian.
\newblock Foldingnet: Point cloud auto-encoder via deep grid deformation.
\newblock In {\em Proceedings of the IEEE Conference on Computer Vision and
  Pattern Recognition}, pages 206--215, 2018.

\bibitem{yi2018gspn}
Li Yi, Wang Zhao, He Wang, Minhyuk Sung, and Leonidas Guibas.
\newblock Gspn: Generative shape proposal network for 3d instance segmentation
  in point cloud.
\newblock {\em arXiv preprint arXiv:1812.03320}, 2018.

\bibitem{zhou2018voxelnet}
Yin Zhou and Oncel Tuzel.
\newblock Voxelnet: End-to-end learning for point cloud based 3d object
  detection.
\newblock In {\em Proceedings of the IEEE Conference on Computer Vision and
  Pattern Recognition}, pages 4490--4499, 2018.

\end{thebibliography}
}

\newpage
\appendix
\section{Appendix}
\label{sec:overview}
This appendix provides additional details on the network architectures and loss functions (Sec.~\ref{sec:arch_details}), more analysis experiment results (Sec.~\ref{sec:more_analysis}), per-category results on ScanNet (Sec.~\ref{sec:percat_scannet}), and finally more visualizations (Sec.~\ref{sec:more_vis}).
% \begin{itemize}
%     \item (must have) More than one vote.
%     \item (must have) Loss functions.
%     \item (must have) Network architectures on \votenet{} and BoxNet.
%     \item (must have) Per-category results on ScanNetV2
%     \item Number of proposals v.s. recall/mAP
%     \item Vote context: SA4,FP1-4.
%     \item Two-stage detectors.
%     \item sampling strategies on vote clustering.
%     \item (optional) instance segmentation.
%     \item (optional) more than one proposal per vote cluster.
% \end{itemize}

\subsection{Details on Architectures and Loss Functions}
\label{sec:arch_details}

\paragraph{\votenet{} architecture details.} As mentioned in the main paper, the \votenet{} architecture composes of a backbone point feature learning network, a voting module and a proposal module.

The backbone network, based on the PointNet++ architecture~\cite{qi2017pointnetplusplus}, has four set abstraction layers and two feature up-sampling layers. The detailed layer parameters are shown in Table~\ref{tab:votingnet_detail}. Each set abstraction (SA) layer has a receptive field specified by a ball-region radius $r$, a MLP network for point feature transform $MLP[c_1,...,c_k]$ where $c_i$ is output channel number of the $i$-th layer in the MLP. The SA layer also subsamples the input point cloud with farthest point sampling to $n$ points. Each SA layer is specified by $(n, r, [c_1,...,c_k])$ as shown in the Table~\ref{tab:votingnet_detail}. Compared to~\cite{qi2017pointnetplusplus}, we also normalize the XYZ scale of points in each local region by the region radius.

Each set feature propagation (FP) layer upsamples the point features by interpolating the features on input points to output points (each output point's feature is weighted average of its three nearest input points' features). It also combines the skip-linked features through a MLP (interpolated features and skip-linked features are concatenated before fed into the MLP). Each FP layer is specified by $[c_1,...,c_k]$ where $c_i$ is the output of the $i$-th layer in the MLP.

The voting module as mentioned in the main paper is a MLP that transforms seeds' features to votes including a XYZ offset and a feature offset. The seed points are outputs of the fp2 layer. The voting module MLP has output sizes of $256,256,259$ for its fully connected layers. The last fully connected layer does not have ReLU or BatchNorm.

The proposal module as mentioned in the main paper is a SA layer followed by another MLP after the max-pooling in each local region. We follow~\cite{qi2018frustum} on how to parameterize the oriented 3D bounding boxes. The layer's output has $5+2NH+4NS+NC$ channels where $NH$ is the number of heading bins (we predict a classification score for each heading bin and a regression offset for each bin -- relative to the bin center and normalized by the bin size), $NS$ is the number of size templates (we predict a classification score for each size template and 3 scale regression offsets for height, width and length) and $NC$ is the number of semantic classes. In SUN RGB-D: $NH = 12, NS = NC = 10$, in ScanNet: $NH = 12, NS = NC = 18$. In the first $5$ channels, the first two are for objectness classification and the rest three are for center regression (relative to the vote cluster center). 

\begin{table*}[]
    \begin{center}
    \begin{tabular}{|c|c|c|c|c|}
    \hline
        layer name & input layer & type & output size &  layer params\\ \hline
        sa1 & raw point cloud & SA & (2048,3+128) & (2048,0.2,[64,64,128]) \\ 
        sa2 & sa1 & SA & (1024,3+256) & (1024,0.4,[128,128,256])\\
        sa3 & sa2 & SA & (512,3+256) & (512,0.8,[128,128,256]) \\
        sa4 & sa3 & SA & (256,3+256) & (256,1.2,[128,128,256]) \\
        fp1 & sa3, sa4 & FP & (512,3+256) & [256,256] \\
        fp2 & sa2, sa3 & FP & (1024,3+256) & [256,256] \\ \hline
    \end{tabular}
    \end{center}
    \caption{Backbone network architecture: layer specifications.}
    \label{tab:votingnet_detail}
\end{table*}

\paragraph{\votenet{} loss function details.}
The network is trained end-to-end with a multi-task loss including a voting loss, an objectness loss, a 3D bounding box estimation loss and a semantic classification loss. We weight the losses such that they are in similar scales with $\lambda_1 = 0.5$, $\lambda_2 = 1$ and $\lambda_3 = 0.1$.

\begin{equation}
    L_{\text{\votenet{}}} = L_{\text{vote-reg}} + \lambda_1 L_{\text{obj-cls}} + \lambda_2 L_{\text{box}} + \lambda_3 L_{\text{sem-cls}}
\end{equation}

Among the losses, the vote regression loss is as defined in the main paper (with L1 distance). For ScanNet we compute the ground truth votes as offset from the mesh vertices of an instances to the centers of the axis-aligned tight bounding boxes of the instances. Note that since the bounding box is not amodal, they can vary in sizes due to scan completeness (e.g. a chair may have a floating bounding box if its leg is not recovered from the reconstruction). For SUN RGB-D since the dataset does not provide instance segmentation annotations but only amodal bounding boxes, we cannot compute a ground truth vote directly (as we don't know which points are on objects). Instead, we consider any point inside an annotated bounding box as an object point (required to vote) and compute its offset to the box center as the ground truth. In cases that a point is in multiple ground truth boxes, we keep a set of up to three ground truth votes, and consider the minimum distance between the predicted vote and any ground truth vote in the set during vote regression on this point.

The objectness loss is just a cross-entropy loss for two classes and the semantic classification loss is also a cross-entropy loss of $NC$ classes.

The box loss follows~\cite{qi2018frustum} (but without the corner loss regularization for simplicity) and is composed of center regression, heading estimation and size estimation sub-losses. In all regression in the box loss we use the robust $L1$-smooth loss. Both the box and semantic losses are only computed on positive vote clusters and normalized by the number of positive clusters. We refer readers to~\cite{qi2018frustum} for more details.

\begin{equation}
\begin{split}
    L_{\text{box}} & = L_{\text{center-reg}} + 0.1 L_{\text{angle-cls}} + L_{\text{angle-reg}} \\
    & + 0.1 L_{\text{size-cls}} + L_{\text{size-reg}}
\end{split}
\end{equation}

One difference though is that, instead of a naive regression loss, we use a \emph{Chamfer loss}~\cite{fan2017point} for $L_{\text{center-reg}}$ (between regressed centers and ground truth box centers). It requires that each positive proposal is close to a ground truth object and each ground truth object center has a proposal near it. The latter part also influences the voting in the sense that it encourages non-object seed points near the object to also vote for the center of the object, which helps further increase contexts in detection.

\paragraph{BoxNet architecture details.}
Our baseline network without voting, BoxNet, shares most parts with the \votenet{}. They share the same backbone architecture. But instead of voting from seeds, the BoxNet directly proposes bounding boxes and classifies object classes from seed points' features. To make the BoxNet and \votenet{} have similar capacity we also include a SA layer for the proposal in BoxNet. However this SA layer takes \emph{seed clusters} instead of \emph{vote clusters} i.e. it samples seed points and then combines neighboring seeds with MLP and max-pooling. This SA layer has exactly the same layer parameters with that in the \votenet{}, followed by the same $MLP_2$.

\paragraph{BoxNet loss function details.}
BoxNet has the same loss function as \votenet{}, except it is not supervised by vote regression. There is also a slight difference in how objectness labels (used to supervise objectness classification) are computed. As seed points (on object surfaces) are often far from object centroids, it no longer works well to use the distances between seed points and object centroids to compute the objectness labels. In BoxNet, we assign positive objectness labels to seed points that are on objects (those belonging to the semantic categories we consider) and negative labels to all the other seed points on clutter (e.g. walls, floors).

\begin{equation}
    L_{\text{BoxNet}} = \lambda_1 L_{\text{obj-cls}} + \lambda_2 L_{\text{box}} + \lambda_3 L_{\text{sem-cls}}
\end{equation}

\begin{table*}[t!]
\begin{center}
\footnotesize
\setlength{\tabcolsep}{3pt}
\begin{tabular}{l|cccccccccccccccccc|c}
\toprule
& cab & bed & chair & sofa & tabl & door & wind & bkshf & pic & cntr & desk & curt & fridg & showr & toil & sink & bath & ofurn & mAP \\ \midrule
3DSIS 5views~\cite{hou20183d} & 19.76 & 69.71 & 66.15 & 71.81 & 36.06 & 30.64 & 10.88 & 27.34 & 0.00 & 10.00 & 46.93 & 14.06 & 53.76 & 35.96 & 87.60 & 42.98 & 84.30 & 16.20 & 40.23 \\
3DSIS Geo~\cite{hou20183d} & 12.75 & 63.14 & 65.98 & 46.33 & 26.91 & 7.95 & 2.79 & 2.30 & 0.00 & 6.92 & 33.34 & 2.47 & 10.42 & 12.17 & 74.51 & 22.87 & 58.66 & 7.05 & 25.36 \\
%\votenet{} (ours) & 29.96 & 82.87 & 78.66 & 76.32 & 53.10 & 29.64 & 23.57 & 33.59 & 2.37 & 36.92 & 57.76 & 33.71 & 33.68 & 45.14 & 86.98 & 37.94 & 79.77 & 19.58 & 46.75 \\
\votenet{} {ours} & 36.27 & 87.92 & 88.71 & 89.62 & 58.77 & 47.32 & 38.10 & 44.62 & 7.83 & 56.13 & 71.69 & 47.23 & 45.37 & 57.13 & 94.94 & 54.70 & 92.11 & 37.20 & 58.65 \\
\bottomrule
\end{tabular}
\end{center}
\caption{3D object detection scores per category on the ScanNetV2 dataset, evaluated with mAP@0.25 IoU.}
\label{tab:perclassscannet025}
\end{table*}

\begin{table*}[t!]
\begin{center}
\footnotesize
\setlength{\tabcolsep}{3.2pt}
\begin{tabular}{l|cccccccccccccccccc|c}
\toprule
& cab & bed & chair & sofa & tabl & door & wind & bkshf & pic & cntr & desk & curt & fridg & showr & toil & sink & bath & ofurn & mAP \\ \midrule
3DSIS 5views~\cite{hou20183d} & 5.73 & 50.28 & 52.59 & 55.43 & 21.96 & 10.88 & 0.00 & 13.18 & 0.00 & 0.00 & 23.62 & 2.61 & 24.54 & 0.82 & 71.79 & 8.94 & 56.40 & 6.87 & 22.53 \\
3DSIS Geo~\cite{hou20183d} & 5.06 & 42.19 & 50.11 & 31.75 & 15.12 & 1.38 & 0.00 & 1.44 & 0.00 & 0.00 & 13.66 & 0.00 & 2.63 & 3.00 & 56.75 & 8.68 & 28.52 & 2.55 & 14.60 \\
%\votenet{} (ours) & 6.42 & 71.26 & 45.14 & 50.19 & 31.88 & 6.15 & 4.04 & 22.32 & 0.08 & 5.14 & 21.32 & 9.36 & 14.27 & 7.75 & 65.56 & 16.50 & 62.86 & 3.53 & 24.65 \\
\votenet{} (ours) & 8.07 & 76.06 & 67.23 & 68.82 & 42.36 & 15.34 & 6.43 & 28.00 & 1.25 & 9.52 & 37.52 & 11.55 & 27.80 & 9.96 & 86.53 & 16.76 & 78.87 & 11.69 & 33.54 \\
\bottomrule
\end{tabular}
\end{center}
\caption{3D object detection scores per category on the ScanNetV2 dataset, evaluated with mAP@0.5 IoU.}
\label{tab:perclassscannet050}
\end{table*}

\subsection{More Analysis Experiments}
\label{sec:more_analysis}

\begin{figure}
    \centering
    \includegraphics[width=\linewidth]{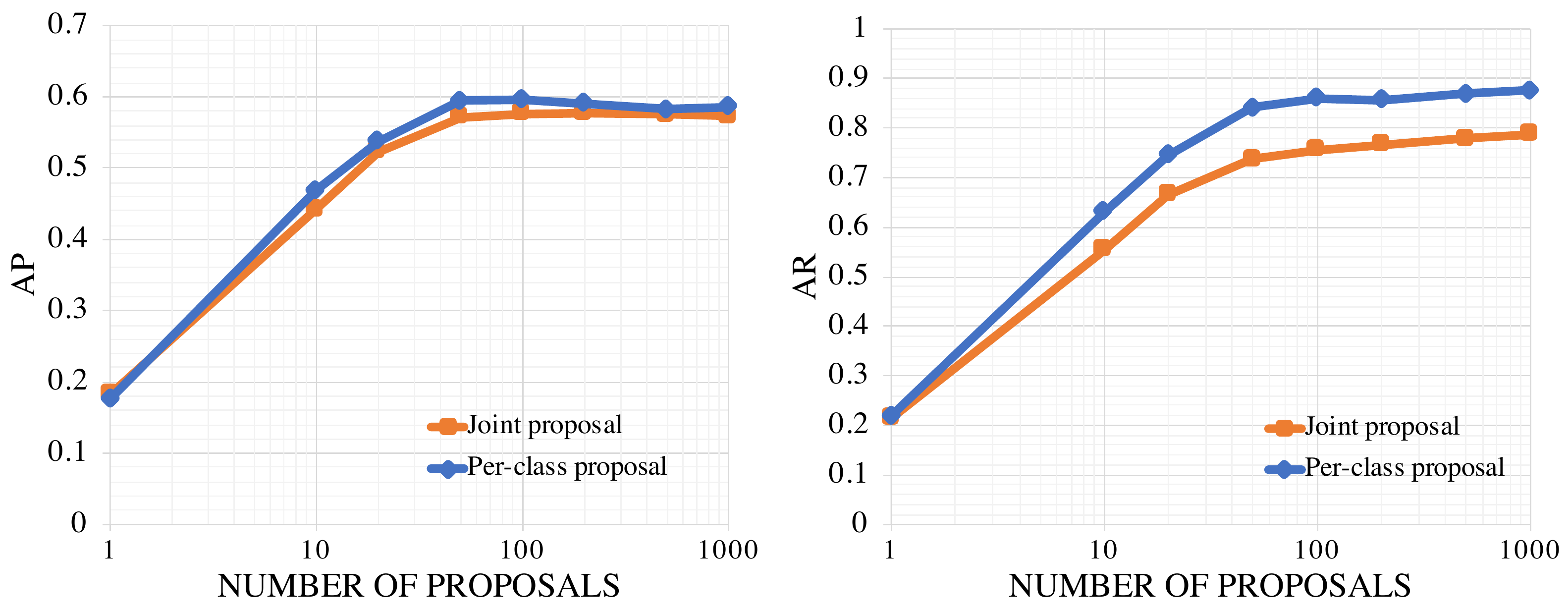}
    \caption{\textbf{Number of proposals per scene v.s. Average Precision (AP) and Average Recall (AR) on SUN RGB-D.} The AP and AR are averaged across the 10 classes. The recall is maximum recall given a fixed number of detection per scene. The ``joint proposal'' means that we assign each proposal to a single class (the class with the highest classification score); The ``per-class proposal'' means that we assign each proposal to all the 10 classes (the objectness score is multipled by the semantic classification probability).}
    \label{fig:ap_ar}
\end{figure}

\paragraph{Average precision and recall plots}
Fig.~\ref{fig:ap_ar} shows how average precision (AP) and average recall (AR) change as we increase the number of proposals. The AP and AR are both averaged across 10 categories on SUN RGB-D. We report two ways of using the proposals: joint and per-class. For the joint proposal we propose $K$ objects' bounding boxes for all the 10 categories, where we consider each proposal as the semantic class it has the largest confidence in, and use their objectness scores to rank them. For the per-class proposal we duplicate the $K$ proposal 10 times thus have $K$ proposals per class where we use the multiplication of semantic probability for that class and the objectness probability to rank them. The latter way of using proposals gives us  a slight improvement on AP and a big boost on AR.

We see that with as few as 10 proposals our \votenet{} can achieve a decent AP of around $45\%$ while having 100 proposals already pushes the AP to above $57\%$. With a thousand proposals, our network can achieve around $78.7\%$ recall with joint proposal and around $87.7\%$ recall with per-class proposal.

\paragraph{Context of voting}
One difference of a deep Hough voting scheme with the traditional Hough voting is that we can take advantage of deep features, which can provide more context knowledge for voting. In Table~\ref{tab:vote_context} we show how features from different levels of the PointNet++ affects detection performance (from SA2 to FP3, the network has increasing contexts for voting). FP3 layer is extended from the FP2 with a MLP of output sizes 256 and 256 with 2048 output points (the same set of XYZ as that output by SA1).

It is surprising to find that voting from even SA2 can achieve reasonable detection results (mAP $51.2\%$) while voting from FP2 achieves the best performance. Having larger context (e.g. FP3) than FP2 does not show further improvements on the performance.

\begin{table}[h]
    \begin{center}
    \begin{tabular}{l|cccccc}
    \toprule
         Seed layer & SA2 & SA3 & SA4 & FP1 & FP2 & FP3 \\ \midrule
         mAP & 51.2 & 56.3 & 55.1 & 56.6 & \textbf{57.7} & 57.1 \\ \bottomrule
    \end{tabular}
    \end{center}
    \caption{\textbf{Effects of seed context for 3D detection.} Evaluation metric is mAP@0.25 on SUN RGB-D.}
    \label{tab:vote_context}
\end{table}

% \paragraph{Two-stage detection}
% Is the \boxnet{} on par or better than \votenet{} with a two-stage detection scheme?

\paragraph{Multiple votes per seed}
% How to support multiple votes per seed and does it help?
In default we just generate one vote per seed since we find that with large enough context there is little need to generate more than one vote to resolve ambiguous cases. However, it is still possible to generate more than one vote with our network architecture. Yet to break the symmetry in multiple vote generation, one has to introduce some bias to different votes to prevent then from pointing to the same place.

In experiments, we find that one vote per seed achieves the best results, as shown in Table~\ref{tab:vote_num}. We ablate by using a vote factor of $3$, where the voting module generates $3$ votes per seed with a MLP layer spec: $[256,256,259*3]$). In computing the vote regression loss on a seed point, we consider the minimum distance between any predicted votes to the ground truth vote (in case of SUN RGB-D where we may have a set of ground truth votes for a seed, we compute the minimum distance among any pair of predicted vote and ground truth vote).

\begin{figure}[t!]
    \centering
    % \vspace{14pt}
    \begin{overpic}
    [trim=0cm 0cm 0cm 0cm,clip,width=\linewidth]{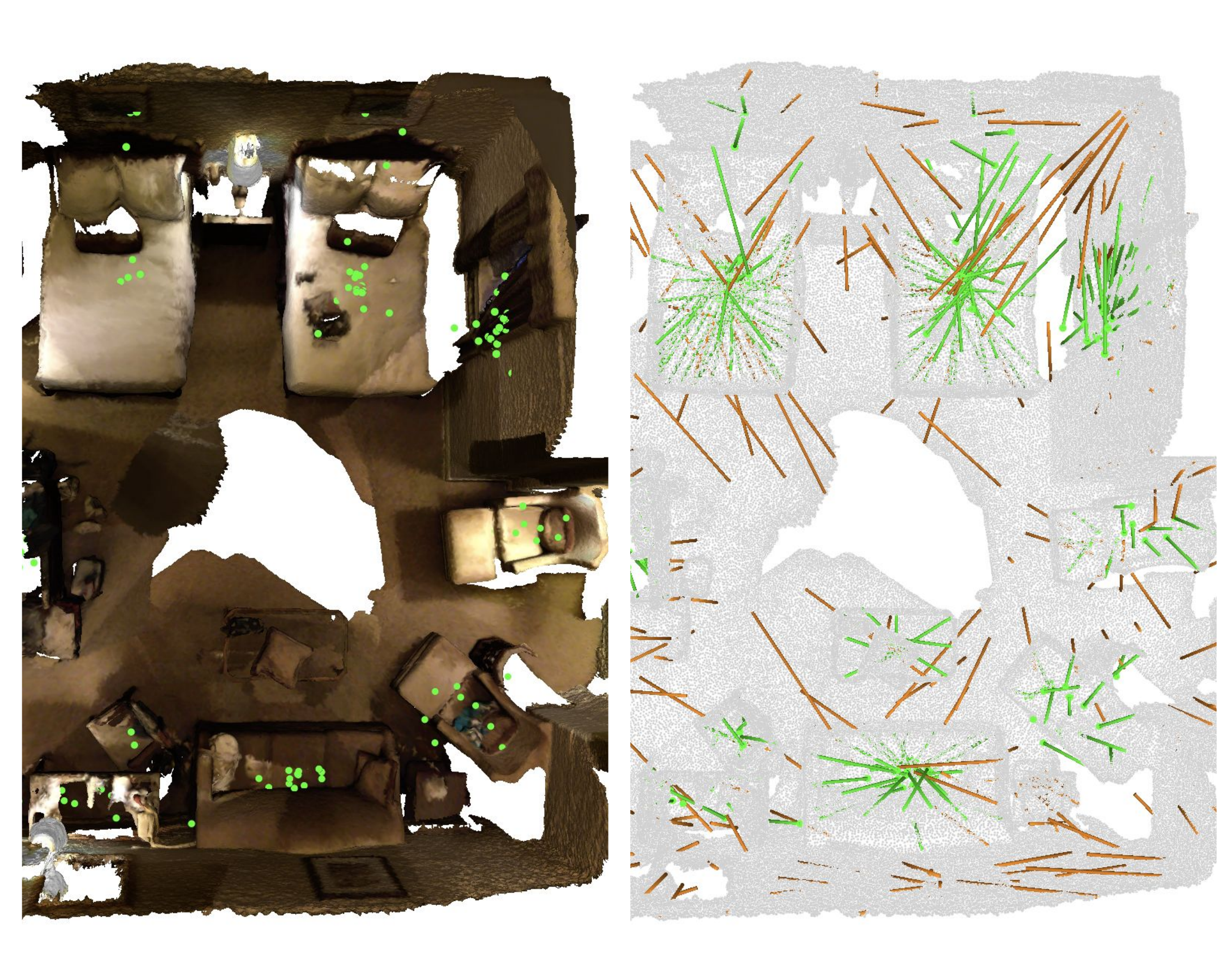}
    % \put(2,43){\small Voting from input point cloud}
    % \put(60,43){\small 3D detection output}
    \end{overpic}
    \caption{\textbf{Vote meeting point.} \emph{Left:} ScanNet scene with votes coming from object points. \emph{Right:} vote offsets from source seed-points to target-votes. Object votes are colored green, and non-object ones are colored red. See how object points from all-parts of the object vote to form a cluster near the center. Non-object points, however, either vote ``nowhere'' and therefore lack structure, or are near object and have gathered enough context to also vote properly. }
    \label{fig:showvotes}
\end{figure}

To break symmetry, we generate $3$ random numbers and inject them to the second last features from the MLP layer. We show results both with and without this procedure which shows no observable difference.  %In both cases, the network with $3$ votes per seed perform worse than the one with 1 vote per seed.

\begin{table}[]
    \begin{center}
    \begin{tabular}{c|c|c|c}
    \toprule
         Vote factor & 1 & 3 & 3 \\ \hline
         Random number & N & N & Y \\ \hline
         mAP & \textbf{57.7} & 55.8 & 55.8 \\ \bottomrule
    \end{tabular}
    \end{center}
    \caption{\textbf{Effects of number of votes per seed.} Evaluation metric is mAP@0.25 on SUN RGB-D. If random number is on, we concatenate a random number to the seed feature before voting, which helps break symmetry in the case of multiple votes per seed.}
    \label{tab:vote_num}
\end{table}

\paragraph{On proposal sampling}
In the proposal step, to generate $K$ proposals from the votes, we need to select $K$ vote clusters. How to select those clusters is a design choice we study here (each cluster is simply a group of votes near a center vote). In Table~\ref{tab:sampling}, we report mAP results on SUN RGB-D with 256 proposals (joint proposal) using cluster sampling strategies of vote FPS, seed FPS and random sampling, where FPS means farthest point sampling. From $1024$ vote clusters, vote FPS samples $K$ clusters based on votes' XYZ. Seed FPS firstly samples on seed XYZ and then finds the votes corresponding to the sampled seeds -- it enables a direct comparison with BoxNet as it uses the same sampling scheme, making the two techniques similar up to the space in which the points are grouped: \votenet{} groups votes according to vote XYZ, while BoxNet groups seeds according to seed XYZ. Random sampling simply selects a random set of $K$ votes and take their neighborhoods for proposal generation. Note that the results from Table~\ref{tab:sampling} are from the same model trained with vote FPS to select proposals.

We can see that while seed FPS gets the best number in mAP, the difference caused by different sampling strategies is small, showing the robustness of our method.

\begin{table}[]
    \begin{center}
    \begin{tabular}{l|c}
    \toprule
        Proposal sampling & mAP \\ \midrule
        Random sampling & 57.5\\ % 0.577005, 0.573915
        Farthest point sampling on votes & 57.2\\ % 0.569167 0.575852
        Farthest point sampling on seeds & \textbf{57.7}\\ 
        \bottomrule
    \end{tabular}
    \end{center}
    \caption{\textbf{Effects of proposal sampling.} Evaluation metric is mAP@0.25 on SUN RGB-D. $256$ proposals are used for all evaluations. Our method is not sensitive to how we choose centers for vote groups/clusters.}
    \label{tab:sampling}
\end{table}

% FPS on seeds means that we choose the indices of votes based on the 3D Euclidean distances of the seeds corresponding to the votes.
% FPS on seeds is the same way as how proposal locations are chosen in the BoxNet, but differently we group local information in the vote $XYZ$ space thus getting a different neighborhood and more effective context.
\paragraph{Effects of the height feature}
In point clouds from indoor scans, point height is a useful feature in recognition. As mentioned in the main paper, we can use $1\%$ of the $Z$ values ($Z$-axis is up-right) of all points from a scan as an approximate as the floor height $z_{\text{floor}}$, and then compute the a point $(x,y,z)$'s height as $z - z_{\text{floor}}$. In Table~\ref{tab:height} we show how this extra height feature affect detection performance. We see that adding the height feature consistently improves performance in both SUN RGB-D and ScanNet.

\begin{table}[]
    \centering
    \begin{tabular}{l|c|c}
    \toprule
         Dataset & with height & without height \\ \midrule
         SUN RGB-D & 57.7 & 57.0 \\ \hline
         ScanNet & 58.6 & 58.1 \\ \bottomrule
    \end{tabular}
    \caption{\textbf{Effects of the height feature.} Evaluation metric is mAP@0.25 on both datasets.}
    \label{tab:height}
\end{table}

\subsection{ScanNet Per-class Evaluation}
\label{sec:percat_scannet}

Table~\ref{tab:perclassscannet025} and Table~\ref{tab:perclassscannet050} report per-class average precision on 18 classes of ScanNetV2 with $0.25$ and $0.5$ box IoU thresholds respectively. Relying on purely geonetric data, our method excels (esp. with mAP@0.25) in detecting objects like bed, chair, table, desk etc. where geometry is a strong cue for recognition; and struggles with objects best recognized by texture and color like pictures.

\subsection{Visualization of Votes}
\label{sec:more_vis}
% \begin{figure*}
%     \centering
%     \includegraphics[width=\linewidth,height=6cm]{./fig/placeholder.pdf}
%     \caption{\todo{Visualizing votes from clutter seeds and object seeds.}}
%     \label{fig:my_label}
% \end{figure*}
Fig.~\ref{fig:showvotes} shows (a subset of) votes predicted from our \votenet{} in a typical ScanNet scene. We clearly see that seed points on objects (bed, sofa etc.) vote to object centers while clutter points vote either to object center as well (if the clutter point is close to the object) or to nowhere due to lack of structure in the clutter area (e.g. a wall).

\end{document}